\documentclass[]{fairmeta}

\usepackage{amsmath,amssymb}
\usepackage{algorithm}
\usepackage{algpseudocode}
\usepackage{colortbl}

\colorlet{myblue}{metabg!7}
\definecolor{mygrey}{HTML}{F1F2F3}

\algrenewcommand\algorithmicrequire{\textbf{Input:}}
\algrenewcommand\algorithmicensure{\textbf{Output:}}
\newcommand{\Phase}[1]{\Statex \hskip-\algorithmicindent \textit{\textbf{#1}}}
\newcommand{\method}{\textsc{TTSP}}

\title{Test-Time Scaling over Perception: Resolving the Grounding Paradox in Thinking with Images}

\author[1]{Zheng Jiang}
\author[1]{Yiming Chen}
\author[2]{Nan He}
\author[1]{Jiahui Chen}
\author[1]{Chaoyang Li}
\author[1]{Houde Qian}
\author[1]{Lifeng Sun}

\affiliation[1]{Tsinghua University, Beijing, China}
\affiliation[2]{Beijing University of Technology, Beijing, China}

\metadata[Code]{\url{https://github.com/jiangz20/TTSP}}
\metadata[Contact]{\email{jz24@mails.tsinghua.edu.cn}, \email{sunlf@tsinghua.edu.cn}}

\abstract{
Recent multimodal large language models (MLLMs) support \emph{Thinking
with Images}, invoking visual tools such as zooming and cropping to
inspect image regions during inference. Yet these systems remain brittle
in fine-grained reasoning: to acquire a decisive detail, a model must
ground its attention on the correct region, but knowing which region is
correct presupposes having already observed that detail. We identify this
circular dependency as the \emph{grounding paradox}, show that grounding
errors are rarely self-corrected within a single trajectory---once a
misleading region is inspected, all subsequent reasoning conditions on
that observation and the error propagates to the final answer---and
observe that because each trajectory constructs its own evidence,
answer-level aggregation discards the very information that
distinguishes trajectories. We propose \textbf{Test-Time Scaling over
Perception (TTSP)}, a closed-loop framework that treats perception as the
unit of scalable inference and allocates compute along two axes:
\emph{Entropy-Gated Perceptual Exploration} samples diverse trajectories
and uses critical-token entropy to withhold evidence the model cannot
commit to, while \emph{Evidence-Guided Iterative Refinement} distills
validated observations into a correctable Evidence Ledger that steers
later rounds to re-inspect unresolved regions. Across high-resolution and
general multimodal benchmarks, TTSP consistently outperforms strong
test-time scaling baselines, while
improving grounding quality with favorable token efficiency.
}

\begin{document}
\thispagestyle{firstheader}
\maketitle
\pagestyle{empty}

\section{Introduction}
\label{sec:intro}
\begin{figure}[t]
  \centering
  \includegraphics[width=0.85\linewidth]{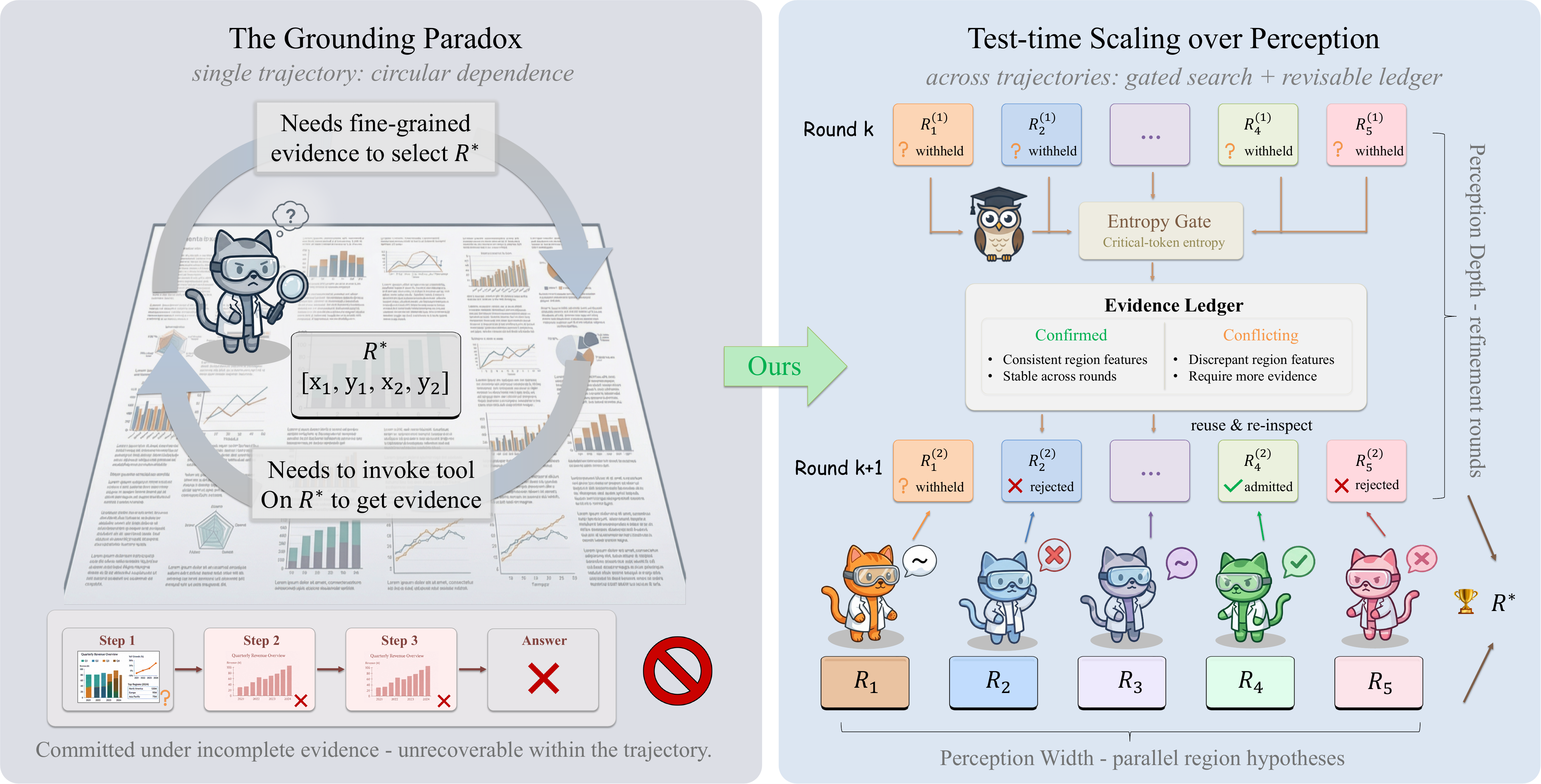}
  \caption{\textbf{The grounding paradox and its resolution.}
  \textbf{Left:} a single trajectory must commit to a region under
  incomplete evidence and inherits any early error. \textbf{Right:}
  \method\ converts the circularity into a gated search across
  trajectories, coupled through a correctable evidence ledger.}
  \label{fig:motivation}
\end{figure}

\emph{Thinking with images} has emerged as a powerful paradigm for
multimodal reasoning: rather than reasoning over a single static view, a
tool-augmented multimodal large language model (MLLM) actively zooms
into, crops, and re-inspects selected image regions during inference,
interleaving these visual operations with its textual reasoning
\citep{zheng2025deepeyes,zhang2025thyme,jiang2026medvr}. This active
regime, however, introduces a structural difficulty absent under fixed
visual input. To acquire a discriminative detail, the model must first
ground its attention on the correct region $R^{*}$; yet knowing which
region is correct presupposes having already observed that detail.
Grounding is a precondition for acquiring evidence, while sufficient
evidence is a precondition for correct grounding. We term this circular
dependence the \textbf{grounding paradox}.
Its consequence is severe within a single trajectory: the model must
commit to a region under incomplete evidence, and every subsequent step
conditions on the resulting observation. Although the tool interface
permits returning to the full image, a trajectory committed to a
misleading region receives no signal that the decisive evidence was
never acquired, and in practice rationalizes an erroneous observation
far more often than it discards one
\citep{dong2025insight,xu2025visulogic}
(Figure~\ref{fig:motivation}, left). Escaping the paradox therefore
calls for alternative perceptual hypotheses, making test-time scaling
the natural place to seek a remedy.

Test-time scaling samples multiple trajectories for the same input and
aggregates their outputs
\citep{zhang2025survey,wang2022SC,li2024ESC,yan2025mur}. Its standard
form, however, rests on a premise that thinking with images weakens:
extra computation is assumed to diversify how \emph{fixed} evidence is
interpreted, so disagreement between answers is purely interpretive and
can be resolved by counting votes. When each trajectory instead decides
where to look, it also constructs its own evidence: two trajectories may inspect disjoint regions and reach different
answers, forming alternative perceptual histories rather than
alternative chains of thought over shared observations
\citep{tan2025reason,li2025latent,li2026RTWI,wang2025hrbench}. This cuts
both ways: a failure that is rarely corrected within a trajectory becomes
a tractable search across trajectories
(Figure~\ref{fig:motivation}, right), yet flat answer-level aggregation,
while still well defined, becomes information-lossy---it weighs an answer
backed by direct inspection of the decisive region no differently from
one backed by an irrelevant crop---shifting the unit that test-time
compute must scale from the answer to the evidence.

Scaling evidence, however, is not merely sampling more trajectories:
broader coverage admits irrelevant crops and confidently stated errors,
and flat voting offers no mechanism to convert disagreement into a
targeted new observation.
Escaping the paradox therefore requires allocating compute along two
complementary axes. Perception width funds parallel region
hypotheses, raising the chance that at least one trajectory acquires the
decisive evidence; perception depth funds successive rounds that
reuse corroborated observations and revisit unresolved claims. The axes
are interdependent: widening exploration helps only if unreliable
observations are filtered before they are shared, and deepening
refinement is safe only if a mistaken observation cannot silently become
a common premise for later rounds. Coupling them thus demands an
explicit record of what has been observed.

We instantiate this principle as \textbf{Test-Time Scaling over
Perception (\method)}, a closed-loop framework that resolves the
grounding paradox by scaling perception along both axes.
\emph{Entropy-Gated Perceptual Exploration} samples diverse tool-use
trajectories and exploits critical-token entropy to withhold evidence the
model itself cannot commit to, broadening coverage without amplifying
noise. \emph{Evidence-Guided Iterative Refinement} maintains a
correctable Evidence Ledger, a dual-tier record separating
confirmed observations from unresolved conflicts, so that later rounds
reuse settled evidence and re-inspect precisely the contested regions.
Since every round pairs fresh exploration with ledger-guided
trajectories, \method\ keeps broadening visual coverage while
progressively resolving uncertain evidence, rather than converging
prematurely on an early consensus.

Our contributions are as follows:
\begin{itemize}
  \item \textbf{Problem.} We formalize the grounding paradox of thinking
  with images, show that self-reinforcing grounding errors are rarely
  corrected within a trajectory but tractable across trajectories, and
  explain why test-time answer scaling over trajectory-dependent evidence
  loses information.
  \item \textbf{Method.} We propose \method, a closed-loop framework that
  scales perception along width and depth via entropy-gated exploration
  and a correctable, dual-tier Evidence Ledger that turns disagreement
  into targeted re-inspection.
  \item \textbf{Performance.} Under matched trajectory budgets, \method\
  consistently surpasses strong test-time scaling baselines across
  backbones and benchmarks, while improving grounding quality with
  favorable token efficiency.
\end{itemize}

\section{Related Work}
\label{sec:related_work}

\paragraph{Test-time scaling.}
Allocating additional computation at inference time can substantially
improve reasoning without updating model parameters
\citep{guo2025deepseek,snell2024scaling,pan2025learning}. Existing
approaches broadly realize this idea through parallel sampling and
consensus---including self-consistency and its adaptive or early-stopping
variants \citep{wang2022SC,li2024ESC,aggarwal2023ASC}---as well as
reranking \citep{weller2025rank1}, verifier-based selection
\citep{chen2024expanding}, and explicit search over candidate solutions
\citep{yan2025mur}. Their common setting holds the input evidence fixed:
additional trajectories explore alternative derivations from the same
observations, making answer-level comparison an informative way to spend
extra compute. This assumption becomes weaker in tool-augmented visual
reasoning. Once each trajectory chooses its own crops and zooms, it also
constructs a distinct perceptual history; disagreement may therefore
reflect different evidence coverage rather than different reasoning over
shared evidence. Flat voting remains well defined, but discards which
regions were inspected and how directly they support an answer. \method\
retains the inference-scaling principle while shifting its primary object
from candidate answers to active evidence acquisition: parallel compute
broadens perceptual coverage, and sequential compute turns unresolved
disagreement into targeted observations.

\paragraph{Thinking with images.}
Multimodal large language models are moving from passive, one-shot image
understanding toward \emph{thinking with images}, in which textual
reasoning is interleaved with visual actions such as cropping, zooming,
drawing, and high-resolution re-inspection
\citep{achiam2023gpt,zheng2025deepeyes,zhang2025thyme,jiang2026medvr}.
Training and inference schemes for multi-step visual interaction have
improved tasks whose decisive evidence is small or initially illegible
\citep{bai2025multi,su2025thinking,fan2025grit,wu2025reinforcing}, and
connect naturally to broader work on visual programs and agentic tool use
\citep{gupta2023visual,luo2025large}. Most of this literature concentrates
on learning \emph{when} to call a tool and \emph{how} to formulate the
next visual operation. The harder question of \emph{where} to look is
often delegated to the same coarse view that made tool use necessary in
the first place. Consequently, an early mislocalized crop can determine
the evidence available to every later step; despite the ability to return
to the original image, long visual chains frequently rationalize the
acquired view rather than discover that the decisive region was missed
\citep{dong2025insight,xu2025visulogic}. We formulate this circular
dependence as the \textbf{grounding paradox}. Unlike methods that optimize
a single tool-use trajectory, \method\ searches across multiple region
hypotheses and preserves the useful observations they uncover for later
rounds.

\paragraph{Reliability and iterative refinement.}
A complementary line improves test-time decisions through confidence-aware
aggregation or best-of-$N$ selection
\citep{taubenfeld2025CISC,kang2025selfcertainty,fu2025deepconf}, while
self-feedback and recursive reasoning revise an initial solution over
multiple passes \citep{madaan2023selfrefine,zhuang2026test}. Recent work
also studies reliability specifically for thinking with images
\citep{li2026RTWI}. These mechanisms are valuable when uncertainty is
primarily attached to a candidate answer or reasoning trace. In active
perception, however, a fluent and confident trace may still be grounded in
an irrelevant region, while repeated refinement may simply reuse the same
incomplete evidence. Moreover, sharing observations across trajectories
introduces an asymmetric risk: one unsupported claim can bias every later
guided trace. \method\ addresses the two levels separately. Critical-token
entropy withholds traces whose key commitments are unstable, but is not
treated as a correctness verifier; cross-trajectory consistency and direct
image support determine what enters a bounded, reversible Evidence Ledger.
Its Confirmed Knowledge tier avoids redundant inspection, whereas Open
Conflicts retain competing claims together with spatial directives for
the next round. This combination couples reliability estimation to
evidence provenance and makes iterative refinement corrective rather than
merely accumulative.

\section{Method}
\label{sec:methods}

\subsection{Problem Setup}
\label{sec:formulation}

A tool-augmented model $\pi_\theta$ receives an image
$I\in\mathbb{R}^{H\times W\times 3}$, a question $Q$, and an optional
candidate set $\mathcal{O}$, and may invoke a crop operator $\Phi(I,b)$ that
re-encodes the sub-image delimited by a box $b=[x_1,y_1,x_2,y_2]$ as a
standalone input. Sampling from
$\pi_\theta(\cdot\mid I,Q,\mathcal{O},\mathcal{E})$ yields a \emph{perception
trace} $\tau=(r_1,b_1,\Phi(I,b_1),\dots,r_L)$ that interleaves reasoning
segments $r_\ell$ with inspected regions $b_\ell$, where $\mathcal{E}$ is an
optional evidence context. We write $a(\tau)$ for the answer read off $r_L$
and $\mathcal{B}(\tau)=\{b_\ell\}$ for the inspected regions.

Let $R^{*}$ be the region whose dedicated inspection is necessary to answer
$Q$ (taken to be unique for exposition), and let
$\mathrm{succ}(\tau)=\mathbf{1}\bigl[\max_{b\in\mathcal{B}(\tau)}
\mathrm{IoU}(b,R^{*})\geq\delta\bigr]$ indicate whether $\tau$ ever grounds
on it; both $R^{*}$ and the tolerance $\delta$ serve evaluation only and are
unavailable at test time. Two properties make the grounding paradox
concrete. First, fine-grained cues become legible only once the region is
inspected on its own \citep{wu2024vstar,zheng2025deepeyes,wang2025hrbench},
yet the first proposal $b_1$ must be made before any such inspection, so
$p=\Pr[\mathrm{succ}(\tau)=1]$ is bounded away from one.
Second, the failure is self-reinforcing. Although the interface allows a
trace to revisit the full image and issue further crops, every segment
after $b_1$ is conditioned on $\Phi(I,b_1)$: the trace carries no signal
that the decisive evidence was never acquired, and conditioning on a
misleading observation biases subsequent proposals toward rationalizing
it rather than abandoning it, a failure mode repeatedly observed in
tool-augmented MLLMs \citep{dong2025insight,xu2025visulogic}. Since recovery within $\tau$ would require the model to spontaneously
discount its own observation history---rare in practice---we treat
grounding failures as \emph{effectively} absorbing at the trajectory
level and seek recovery through fresh perceptual commitments.

Under stochastic decoding and without shared context, $K$ traces are
conditionally i.i.d.\ given $(I,Q)$, so the probability that some trace
grounds correctly is
\begin{equation}
\label{eq:coverage}
  P_{\mathrm{cov}}(K)=1-(1-p)^{K}\;\longrightarrow\;1
  \qquad\text{whenever } p>0 .
\end{equation}
Equation~\eqref{eq:coverage} is an oracle ceiling, attainable only by a
selector that knows $\mathrm{succ}(\tau)$; closing the gap to it
decomposes our problem into three sub-problems, addressed in turn below.
\textbf{Coverage} asks for hypotheses diverse enough that some trace
inspects $R^{*}$ (Section~\ref{sec:sampling}); \textbf{selection} asks
which traces to trust without access to $R^{*}$
(Section~\ref{sec:gating}); \textbf{utilization} asks how acquired
evidence propagates so that later traces inherit rather than rediscover
it (Section~\ref{sec:refinement}).

\subsection{Overview}
\label{sec:overview}

\begin{figure*}[t]
  \centering
  \includegraphics[width=\linewidth]{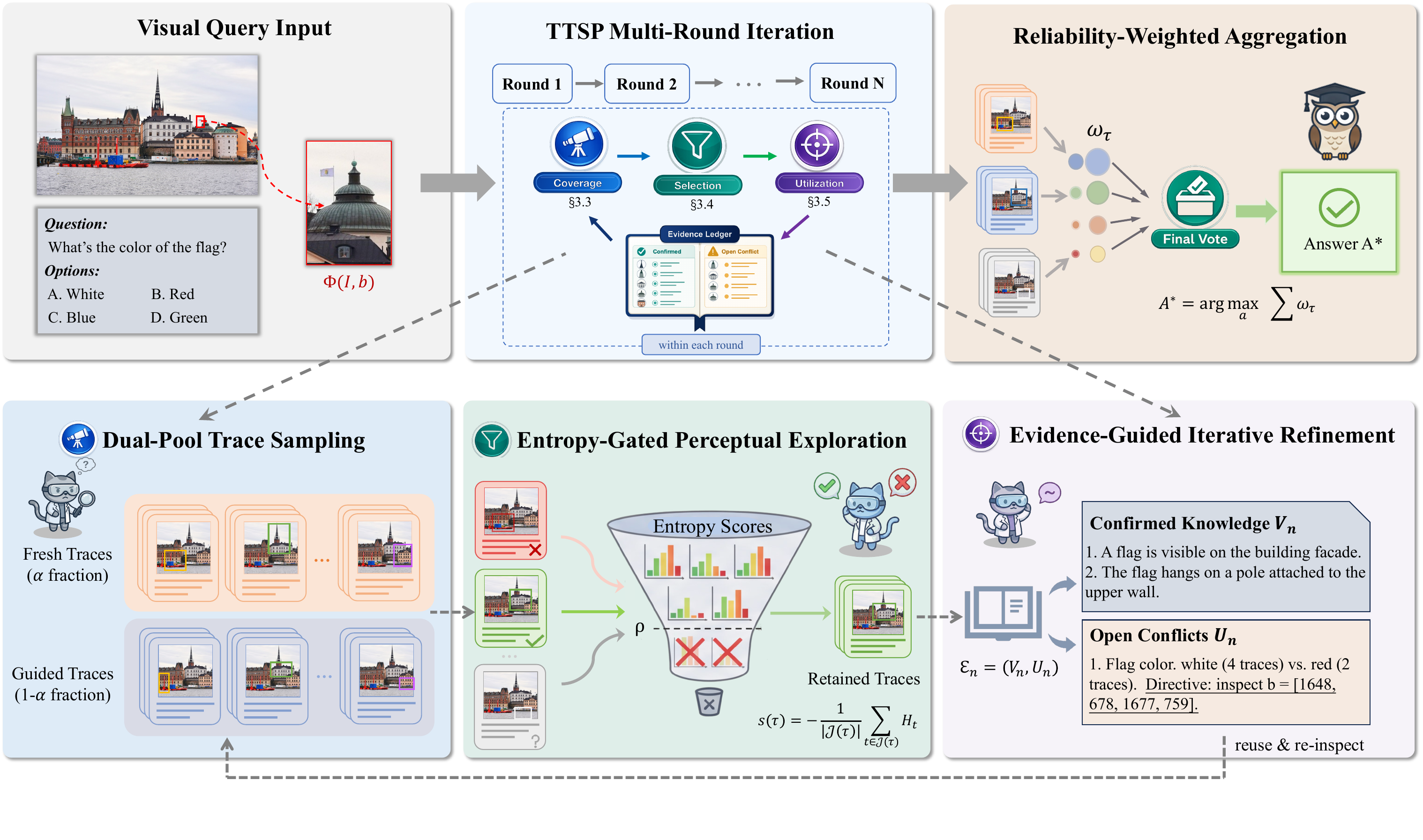}
  \caption{\textbf{Overview of \method.} Each round draws fresh and guided
  traces (\emph{coverage}), gates them on critical-token entropy
  (\emph{selection}), and refines an Evidence Ledger whose spatial
  directives redirect the next round (\emph{utilization}); the answer is a
  reliability-weighted vote over all retained traces.}
  \label{fig:pipeline}
\end{figure*}

\method\ allocates compute along width $K$, the number of hypotheses per
round, and depth $N$, the number of rounds, maintaining two cross-round
states: an evidence context $\mathcal{E}_n$, initialized empty, and a
pool $\mathcal{F}$ of retained traces. Each round advances all three sub-problems in one pass
(Figure~\ref{fig:pipeline}): it draws $K$ traces split between a pool
that ignores $\mathcal{E}_{n-1}$ and one conditioned on it
(\emph{coverage}), scores each trace by the predictive entropy at its
highest-uncertainty decisions and withholds the least reliable fraction
(\emph{selection}), and distills the survivors $\mathcal{F}_n$ into
$\mathcal{E}_n$, a dual-tier Evidence Ledger separating confirmed
observations from open conflicts, injected into the next round
(\emph{utilization}). The final answer is a reliability-weighted vote over the traces retained
in \emph{all} rounds (Algorithm~\ref{alg:ttsp},
Appendix~\ref{app:algorithm}). The two axes are coupled by design: widening
exploration is safe because selection keeps unreliable observations out of
the shared record, and deepening refinement is safe because the ledger is
correctable, so an error cannot silently become a premise for later rounds.

\subsection{Dual-Pool Trace Sampling}
\label{sec:sampling}

Coverage requires that the $K$ traces of a round inspect different
regions, which stochastic decoding at temperature $T_{\mathrm{gen}}>0$
supplies. Once a ledger exists, however, the conditioning context becomes a
design choice. Conditioning every trace on $\mathcal{E}_{n-1}$ would maximize
exploitation but invite confirmation bias, since the traces most likely to
overturn an erroneous entry are those that do not read it first.
\method\ therefore splits the per-round budget into two pools. A fraction
$\alpha$ of \emph{fresh} traces are drawn from
$\pi_\theta(\cdot\mid I,Q,\mathcal{O},\varnothing)$ and preserve an unbiased
exploration channel, keeping the independence behind Eq.~\eqref{eq:coverage}
intact so that coverage keeps growing with the total number of rounds. The
remaining $K-\lceil\alpha K\rceil$ \emph{guided} traces are drawn from
$\pi_\theta(\cdot\mid I,Q,\mathcal{O},\mathcal{E}_{n-1})$ and reuse settled
observations, concentrating crops on contested regions. The split instantiates an exploration--exploitation trade-off over
perceptual hypotheses, controlled by the single knob $\alpha$.

\subsection{Entropy-Gated Perceptual Exploration}
\label{sec:gating}

Broader coverage is a mixed blessing, because the same sampling that
occasionally finds $R^{*}$ also produces irrelevant crops and confidently
stated hallucinations. As $R^{*}$ is unavailable at test time, selection must
rest on a signal intrinsic to generation. Our first core component turns raw
sampling into gated exploration by scoring each trace at the few positions
where it actually commits, and by withholding those whose commitments were
made under high uncertainty.

\paragraph{Critical-token entropy.}
At step $t$ we renormalize the $k_{\mathrm{top}}$ largest token probabilities
returned by the decoder into $\hat p^{(t)}_j$ and take the truncated entropy
$H_t=-\sum_{j\leq k_{\mathrm{top}}}\hat p^{(t)}_j\log\hat p^{(t)}_j$,
obtained at no extra forward pass. Averaging $H_t$ over all positions is
length-sensitive and poorly discriminative, since most tokens are routine
linguistic content emitted with near-zero uncertainty. What determines
whether a trace succeeds are its few highest-uncertainty commitments, namely
choosing box coordinates, reading an ambiguous local cue, and settling on an
answer. We score a trace over $\mathcal{J}(\tau)$, its $k'$ largest-entropy
positions (all positions if shorter):
\begin{equation}
\label{eq:score}
  s(\tau)=-\frac{1}{|\mathcal{J}(\tau)|}\sum_{t\in\mathcal{J}(\tau)}H_t ,
\end{equation}
Since $|\mathcal{J}(\tau)|\leq k'$ regardless of trace length, this
fixed-order-statistic average is comparable across lengths and avoids
sequence-average length bias.

\paragraph{Relative gating.}
Traces are ranked by $s$ within each round and the lowest fraction $\rho$ is
withheld, leaving $\mathcal{F}_n$ with
$|\mathcal{F}_n|=\lceil(1-\rho)K\rceil$ retained traces. Gating is relative
rather than threshold-based, so it requires no per-dataset or per-backbone
calibration of entropy scales. We withhold rather than down-weight because
the two consumers of $\mathcal{F}_n$ have asymmetric error costs. A weak
trace contributes one small term to the final vote, whereas a single
hallucinated observation entering the ledger biases every later round.
Entropy is moreover an intra-trace criterion, complementary to the
inter-trace consensus rules of Section~\ref{sec:refinement}. The former
screens out unstable reasoning that agreement cannot detect, whereas the
latter screens out confident errors that entropy cannot detect.

\subsection{Evidence-Guided Iterative Refinement}
\label{sec:refinement}

Selection decides what to keep, but keeping evidence is not yet using it. Our
second core component converts the retained traces of each round into a
persistent state that both spares later rounds from rediscovering settled
facts and tells them where to look next.

\paragraph{Evidence Ledger.}
The state is a dual-tier record $\mathcal{E}_n=(\mathcal{V}_n,\mathcal{U}_n)$.
Each entry of \emph{Confirmed Knowledge} $\mathcal{V}_n$ pairs a concise
claim with the region supporting it and is treated later as established. Each
\emph{Open Conflict} in $\mathcal{U}_n$ pairs the competing claims with a
\emph{spatial directive} naming the region whose inspection would adjudicate
them. The directive makes the ledger actionable by acquiring new evidence
rather than recounting votes over old evidence. The tiers assign asymmetric
value: confirmations save computation, while conflicts identify observations
that could change the answer.

\paragraph{Update rules.}
At the end of each non-terminal round, $\mathcal{E}_n$ is obtained from
$\mathcal{E}_{n-1}$ and $\mathcal{F}_n$ by four rules. \textsc{Retain}
carries forward uncontradicted entries of $\mathcal{V}_{n-1}$, and
\textsc{Demote} moves contradicted entries into $\mathcal{U}_n$ together with
the contradicting claim. \textsc{Promote} moves conflicts resolved by
sufficient cross-trace consensus into $\mathcal{V}_n$. \textsc{Confirm}
admits a new claim to $\mathcal{V}_n$ only when it is consistent across
independent retained traces, directly supported by the image, compatible with
$\mathcal{V}_{n-1}$, and relevant to $Q$, and records it in $\mathcal{U}_n$
with a directive otherwise. \textsc{Demote} is what distinguishes the ledger
from a monotone memory. Under monotone accumulation an early false
confirmation would itself be absorbing, reinstating at the level of the
shared record the failure mode that motivated leaving a single trajectory.
The strict \textsc{Confirm} criteria favor precision: omissions cost one
re-inspection round, whereas false confirmations corrupt subsequent
refinement.

\paragraph{Implementation and injection.}
The update is a single greedy-decoding call conditioned on the image, the
question, $\mathcal{E}_{n-1}$, and the highest-scoring subset
$\widetilde{\mathcal{F}}_n\subseteq\mathcal{F}_n$, which bounds context
length. Unlike trace generation, it is deterministic by design, since its
role is to summarize evidence rather than to add hypotheses. We cap
$|\mathcal{V}_n|\leq v_{\max}$ and $|\mathcal{U}_n|\leq u_{\max}$ to bound
prompt overhead. For $n>1$, $\mathcal{E}_{n-1}$ is placed in the system
prompt of the guided pool only. Confirmed Knowledge is marked as verified and
not to be re-checked, which frees crops for unexplored regions, while Open
Conflicts are marked as prioritized targets. Guided traces increasingly focus
on adjudication, whereas fresh traces sample independently of the ledger.

\subsection{Reliability-Weighted Aggregation}
\label{sec:aggregation}

Let $\mathcal{F}=\bigcup_{n=1}^{N}\mathcal{F}_n$. Answers are matched against
$\mathcal{O}$ when options are given, and otherwise mapped to equivalence
classes by standard normalization, with $a(\tau)\equiv a$ denoting class
membership. Reusing the score of Eq.~\eqref{eq:score}, each retained trace
votes with weight $w(\tau)=\exp\bigl(s(\tau)/\gamma\bigr)$, so that
\begin{equation}
\label{eq:voting}
  A^{*}=\arg\max_{a}\sum_{\tau\in\mathcal{F}:\,a(\tau)\equiv a} w(\tau),
\end{equation}
which interpolates between majority voting ($\gamma\to\infty$) and
best-of-$N$ selection ($\gamma\to0$). Depending on the weights only through
their ratios, Eq.~\eqref{eq:voting} is invariant to the additive offset
introduced by truncating the entropy in Eq.~\eqref{eq:score}. Aggregating
over all rounds rather than the last one preserves evidence found early that
a purely sequential scheme would discard.
The readout is moreover deliberately decoupled from the ledger: the ledger
steers where later rounds look, but holds no veto over the final decision.
This asymmetry is intentional. A trace whose claim is later contradicted
contributes at most one discounted vote, an error that is bounded and
diluted as $\mathcal{F}$ grows; granting the ledger the power to exclude
traces would instead make a single erroneous ledger entry a point of
failure for the entire prediction, reinstating exactly the absorbing-error
structure that \method\ is designed to avoid.

\section{Experiments}
\label{sec:experiments}

\begin{table*}[t]
\centering
\caption{\textbf{High-resolution and fine-grained visual reasoning.} All
test-time scaling methods within a backbone block use the same trajectory
budget as \method; reference systems are not budget-matched. Best result
per column within each block is in bold.}
\label{tab:main_result}
\small
\setlength{\tabcolsep}{5pt}
\resizebox{\linewidth}{!}{%
\begin{tabular}{l *{15}{c}}
\toprule
& \multicolumn{3}{c}{V$^{*}$ Bench}
& \multicolumn{3}{c}{HR-Bench-4K}
& \multicolumn{3}{c}{HR-Bench-8K}
& \multicolumn{3}{c}{TreeBench}
& \multicolumn{3}{c}{MME-RealWorld-Lite}\\
\cmidrule(lr){2-4} \cmidrule(lr){5-7} \cmidrule(lr){8-10}
\cmidrule(lr){11-13} \cmidrule(lr){14-16}
Method & Attr. & Spat. & Ovr.
& FSP & FCP & Ovr.
& FSP & FCP & Ovr.
& Perc. & Reas. & Ovr.
& Perc. & Reas. & Ovr.\\
\midrule
\multicolumn{16}{c}{\textit{Reference systems, not budget-matched}} \\
\midrule
GPT-4o & 72.2 & 60.5 & 67.5 & 66.8 & 63.3 & 65.0 & 60.8 & 58.5 & 59.6 & 61.7 & 38.3 & 46.9 & 54.4 & 48.3 & 52.0\\
Thyme & 83.5 & 80.3 & 82.2 & 91.0 & 63.0 & 77.0 & 86.5 & 57.5 & 72.0 & 57.0 & 28.9 & 39.3 & 59.1 & 49.1 & 55.2\\
DeepEyes & 91.3 & 88.2 & 90.1 & 91.3 & 59.0 & 75.1 & 86.8 & 58.5 & 72.6 & 57.7 & 29.3 & 39.8 & 55.4 & 46.8 & 52.1\\
Pixel-Reasoner & 83.5 & 76.3 & 80.6 & 86.0 & 60.3 & 72.9 & 80.0 & 54.3 & 66.9 & 54.3 & 30.1 & 39.0 & 53.1 & 45.6 & 50.1\\
\midrule
\rowcolor{myblue}
\multicolumn{16}{c}{\textit{Qwen3-VL-4B-Instruct}} \\
\midrule
Base & 91.3 & 89.5 & 90.6 & 90.0 & 70.0 & 80.0 & 89.8 & 64.0 & 76.9 & 60.4 & 36.3 & 45.2 & 53.4 & 44.8 & 50.0\\
SC & 92.2 & 90.8 & 91.6 & 95.0 & 72.3 & 83.7 & 91.5 & 73.5 & 82.5 & 64.4 & 37.9 & 47.7 & 55.5 & 47.0 & 52.2\\
CISC & 93.0 & 90.8 & 92.1 & 94.8 & 72.3 & 83.6 & 91.5 & 73.3 & 82.4 & 63.8 & 37.5 & 47.2 & 55.7 & 46.6 & 52.1\\
TRT & 92.2 & 90.8 & 91.6 & 95.0 & 72.3 & 83.7 & 91.5 & 73.3 & 82.4 & 63.8 & 37.5 & 47.2 & 55.5 & 47.0 & 52.1\\
Self-Ref. & 91.3 & 89.5 & 90.6 & 95.3 & 72.5 & 83.9 & 91.3 & 73.5 & 82.4 & 64.4 & 37.9 & 47.7 & 55.4 & 46.6 & 52.0\\
Self-Cer. & 91.3 & 89.5 & 90.6 & 93.0 & 72.3 & 82.7 & 91.5 & 73.5 & 82.5 & 63.7 & 37.5 & 47.2 & 55.5 & 46.8 & 52.1\\
DeepConf & 93.0 & \textbf{92.1} & 92.6 & 95.3 & 72.5 & 83.9 & 92.8 & 73.8 & 83.3 & 65.8 & 37.9 & 48.2 & 55.9 & 47.1 & 52.4\\
RTWI & 93.9 & 90.8 & 92.7 & 95.5 & 73.0 & 84.3 & 93.0 & 73.8 & 83.4 & 66.4 & 39.0 & 49.1 & 56.6 & 47.2 & 52.9\\
\rowcolor{mygrey}
\textbf{\method} & \textbf{94.8} & \textbf{92.1} & \textbf{93.7} & \textbf{97.3} & \textbf{74.3} & \textbf{85.8} & \textbf{93.5} & \textbf{74.5} & \textbf{84.0} & \textbf{70.5} & \textbf{39.4} & \textbf{50.9} & \textbf{57.7} & \textbf{49.0} & \textbf{54.3}\\
\midrule
\rowcolor{myblue}
\multicolumn{16}{c}{\textit{Qwen3-VL-8B-Instruct}} \\
\midrule
Base & 92.2 & 89.5 & 91.1 & 93.0 & 69.0 & 81.0 & 89.8 & 71.3 & 80.6 & 67.1 & 34.4 & 46.4 & 52.9 & 46.2 & 50.2\\
SC & 93.0 & 90.8 & 92.1 & 95.8 & 76.8 & 86.3 & 93.0 & 74.8 & 83.9 & 69.8 & 37.1 & 49.1 & 56.6 & 49.2 & 53.7\\
CISC & 93.9 & 92.1 & 93.2 & 95.5 & 75.8 & 85.7 & 91.0 & 75.8 & 83.4 & 67.8 & 36.7 & 48.2 & 56.8 & 49.1 & 53.8\\
TRT & 92.2 & 90.8 & 91.6 & 95.3 & 77.0 & 86.2 & 93.0 & 74.8 & 83.9 & 69.1 & 37.1 & 48.9 & 56.8 & 49.2 & 53.8\\
Self-Ref. & 92.2 & 90.8 & 91.6 & 93.8 & 77.0 & 85.4 & 93.0 & 74.8 & 83.9 & 69.1 & 36.7 & 48.6 & 56.7 & 49.4 & 53.8\\
Self-Cer. & 93.0 & 89.5 & 91.6 & 95.3 & 75.8 & 85.6 & 88.8 & 75.3 & 82.1 & 67.8 & 36.7 & 48.2 & 57.1 & 49.4 & 54.0\\
DeepConf & 93.9 & 92.1 & 93.2 & 96.0 & 76.8 & 86.4 & 93.5 & 75.5 & 84.5 & 70.4 & 37.9 & 49.9 & 57.2 & 49.7 & 54.3\\
RTWI & 94.8 & 92.1 & 93.7 & 95.8 & 77.0 & 86.4 & 94.5 & 77.0 & 85.8 & 71.1 & 37.9 & 50.1 & 57.8 & 50.1 & 54.8\\
\rowcolor{mygrey}
\textbf{\method} & \textbf{95.7} & \textbf{93.4} & \textbf{94.8} & \textbf{96.8} & \textbf{79.8} & \textbf{88.3} & \textbf{95.0} & \textbf{77.8} & \textbf{86.4} & \textbf{73.2} & \textbf{39.4} & \textbf{51.8} & \textbf{59.0} & \textbf{52.7} & \textbf{56.5}\\
\bottomrule
\end{tabular}
}
\end{table*}

\subsection{Experimental Setup}
\label{sec:setup}

\paragraph{Benchmarks.}
We evaluate two regimes: high-resolution fine-grained reasoning, where the
decisive evidence occupies a small fraction of the image, on V$^{*}$ Bench
\citep{wu2024vstar}, HR-Bench-4K/8K \citep{wang2025hrbench}, TreeBench
\citep{wang2026TreeVGR}, and MME-RealWorld-Lite \citep{zhang2024mme}; and
general multimodal reasoning, where perception is usually not the
bottleneck, on MMStar \citep{chen2024mmstar}, MMBench \citep{liu2024mmbench},
MathVision \citep{wang2024mathvision}, LogicVista
\citep{xiao2024logicvista}, and MathVista \citep{lu2024mathvista}. Results
on VisualProbe are reported in Appendix~\ref{sec:visualprobe}.

\paragraph{Baselines.}
\emph{Reference systems}, namely GPT-4o \citep{achiam2023gpt}, Thyme
\citep{zhang2025thyme}, DeepEyes \citep{zheng2025deepeyes}, and
Pixel-Reasoner \citep{wang2025pixel}, use different backbones and budgets
and only situate benchmark difficulty. \emph{Test-time scaling methods},
namely SC \citep{wang2022SC}, CISC \citep{taubenfeld2025CISC}, TRT
\citep{zhuang2026test}, Self-Refine \citep{madaan2023selfrefine},
Self-Certainty \citep{kang2025selfcertainty}, DeepConf
\citep{fu2025deepconf}, and RTWI \citep{li2026RTWI}, share our backbone and
trajectory budget; RTWI, the only baseline combining \emph{Thinking with
Images} with inference-time scaling, is our primary reference.

\paragraph{Implementation and budget protocol.}
We instantiate \method\ on Qwen3-VL \citep{bai2025qwen3} 4B and 8B Instruct
with $N=4$, $K=8$, $\alpha=0.4$, and $\rho=0.4$ throughout. All test-time
scaling baselines are matched on the total number of sampled trajectories;
the $N-1$ greedy ledger updates of \method\ are a small surplus outside this
count and are included in the token accounting of
Section~\ref{sec:efficiency}. More experimental details appear in Appendices~\ref{app:setup_details},
\ref{sec:generation_hyperparameters}, and~\ref{sec:prompts}.

\subsection{Main Results}
\label{sec:results}

\paragraph{Fine-grained perception.}
Table~\ref{tab:main_result} shows that \method\ attains the best overall
score on every high-resolution benchmark with both backbones. Two patterns stand out. First, the answer-level baselines (SC, CISC,
TRT, Self-Refine, Self-Certainty) are tightly clustered---the empirical
signature of the argument in Section~\ref{sec:formulation} that
redistributing votes cannot recover evidence no trajectory acquired;
the two that do separate, DeepConf and RTWI, rely on exactly the
ingredients \method\ combines and extends with a correctable shared
record. Second, the margin over the strongest baseline
is widest on benchmarks whose questions hinge on locating a small region
and narrowest near saturation, precisely what the grounding paradox
predicts.

\paragraph{Perception versus reasoning.}
TreeBench and MME-RealWorld-Lite separate perception from reasoning, and
\method\ improves both splits rather than the perception split alone, as a
method that merely retrieved better crops would. The category-level
breakdowns in Appendices~\ref{sec:tree_detailed}
and~\ref{sec:mme_detailed} localize the gains to categories that require
reading a small or distant region, with the reasoning categories that
depend on those readings following suit.

\paragraph{General multimodal reasoning.}
Table~\ref{tab:general_results} shows that \method\ ranks first on all
five general suites, where perception is usually not the bottleneck: the
entropy gate withholds unstable trajectories in any domain, and the ledger
converts disagreement into a specific target to re-examine even when that
target is a diagram or a formula. Scaling perception therefore does not
trade competence for fine-grained accuracy.

\begin{table}[t]
\centering
\begin{minipage}[t]{0.56\linewidth}
\vspace{0pt}
\caption{\textbf{General reasoning} with Qwen3-VL-4B-Instruct.}
\label{tab:general_results}
\centering
\scriptsize
\setlength{\tabcolsep}{3pt}
\resizebox{\linewidth}{!}{%
\begin{tabular}{lccccc}
\toprule
Method & MMStar & MMBench & MathVision & LogicVista & MathVista \\
\midrule
Base      & 62.8 & 81.2 & 21.4 & 47.9 & 68.9 \\
SC        & 67.6 & 82.5 & 24.0 & 54.6 & 73.2 \\
CISC      & 68.7 & 82.5 & 24.0 & 53.4 & 73.3 \\
Self-Cer. & 67.6 & 82.4 & 24.0 & 54.8 & 73.2 \\
DeepConf  & 68.7 & 83.2 & 22.7 & 56.0 & 73.7 \\
RTWI      & 70.7 & 83.2 & 26.1 & 58.5 & 73.7 \\
\rowcolor{mygrey}
\textbf{\method} & \textbf{73.1} & \textbf{83.7} & \textbf{28.0} & \textbf{59.8}
& \textbf{77.2}\\
\bottomrule
\end{tabular}%
}
\end{minipage}
\hfill
\begin{minipage}[t]{0.43\linewidth}
\vspace{0pt}
\caption{\textbf{Ablation study} with Qwen3-VL-8B-Instruct.}
\label{tab:ablation_study}
\centering
\scriptsize
\setlength{\tabcolsep}{2.5pt}
\resizebox{\linewidth}{!}{%
\begin{tabular}{lccccccc}
\toprule
& \multicolumn{2}{c}{V$^{*}$ Bench} & \multicolumn{2}{c}{HR-Bench-4K}
& \multicolumn{2}{c}{HR-Bench-8K} & \\
\cmidrule(lr){2-3} \cmidrule(lr){4-5} \cmidrule(lr){6-7}
Variant & Attr. & Spat. & FSP & FCP & FSP & FCP & Avg. \\
\midrule
SC & 93.0 & 90.8 & 95.8 & 76.8 & 93.0 & 74.8 & 87.4 \\
\midrule
w/o EG & 93.0 & \textbf{93.4} & 96.3 & 78.0 & \textbf{95.0} & 77.5 & 88.9 \\
w/o EL & 93.0 & 90.8 & 96.0 & 77.0 & 93.5 & 75.8 & 87.7 \\
w/o WA & 94.8 & \textbf{93.4} & 96.3 & 78.3 & 94.5 & \textbf{77.8} & 89.2 \\
\midrule
\rowcolor{mygrey}
\textbf{\method} & \textbf{95.7} & \textbf{93.4} & \textbf{96.8} & \textbf{79.8}
& \textbf{95.0} & \textbf{77.8} & \textbf{89.8} \\
\bottomrule
\end{tabular}%
}
\end{minipage}
\end{table}
\FloatBarrier

\subsection{Ablation Study}
\label{sec:ablation_study}

\emph{w/o EG} disables the entropy gate ($\rho=0$), \emph{w/o EL}
removes the Evidence Ledger ($\alpha=1$), and \emph{w/o WA} replaces
weighted aggregation with an unweighted vote ($\gamma\!\to\!\infty$).
We read Table~\ref{tab:ablation_study} through the ordering of the
averages, since individual columns tie on saturated subsets. Each variant
disables one mechanism by setting its knob to a degenerate value, which
also perturbs downstream quantities ($\rho=0$ enlarges the retained
pool; $\alpha=1$ removes guided conditioning); the table is thus an
ordering of mechanism importance under a fixed budget rather than an
exact per-module decomposition. Removing the
Evidence Ledger is most damaging, falling back close to plain
self-consistency: without a shared record every round restarts from the
same state and re-derives settled observations. Disabling the entropy gate
is second most damaging and fails differently: coverage is unchanged, but
unreliable observations now enter the ledger and propagate as apparently
settled premises, so the gate is what makes sharing evidence safe.
Replacing the weighted vote costs the least, since the reliability
signal has already done its work at the gate. The variants thus support
the intended division of labor: the gate protects the ledger, the
ledger drives iterative refinement, and the weighted readout extracts
the answer.


\begin{figure*}[htbp]
  \centering
  \includegraphics[width=\linewidth]{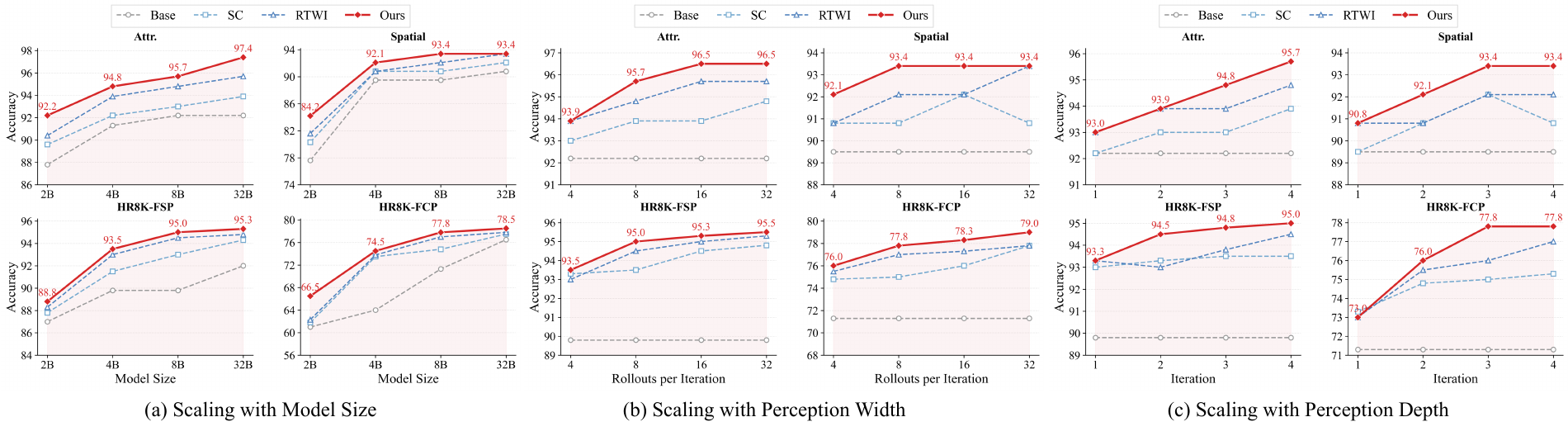}
  \caption{\textbf{Scaling behavior of \method} along (a) backbone size,
  (b) perception width $K$, and (c) perception depth $N$.}
  \label{fig:scaling}
\end{figure*}

\subsection{Analysis}
\label{sec:analysis}

\begin{figure*}[t]
  \begin{minipage}[t]{0.63\linewidth}
    \vspace{0pt}
    \centering
    \includegraphics[width=\linewidth]{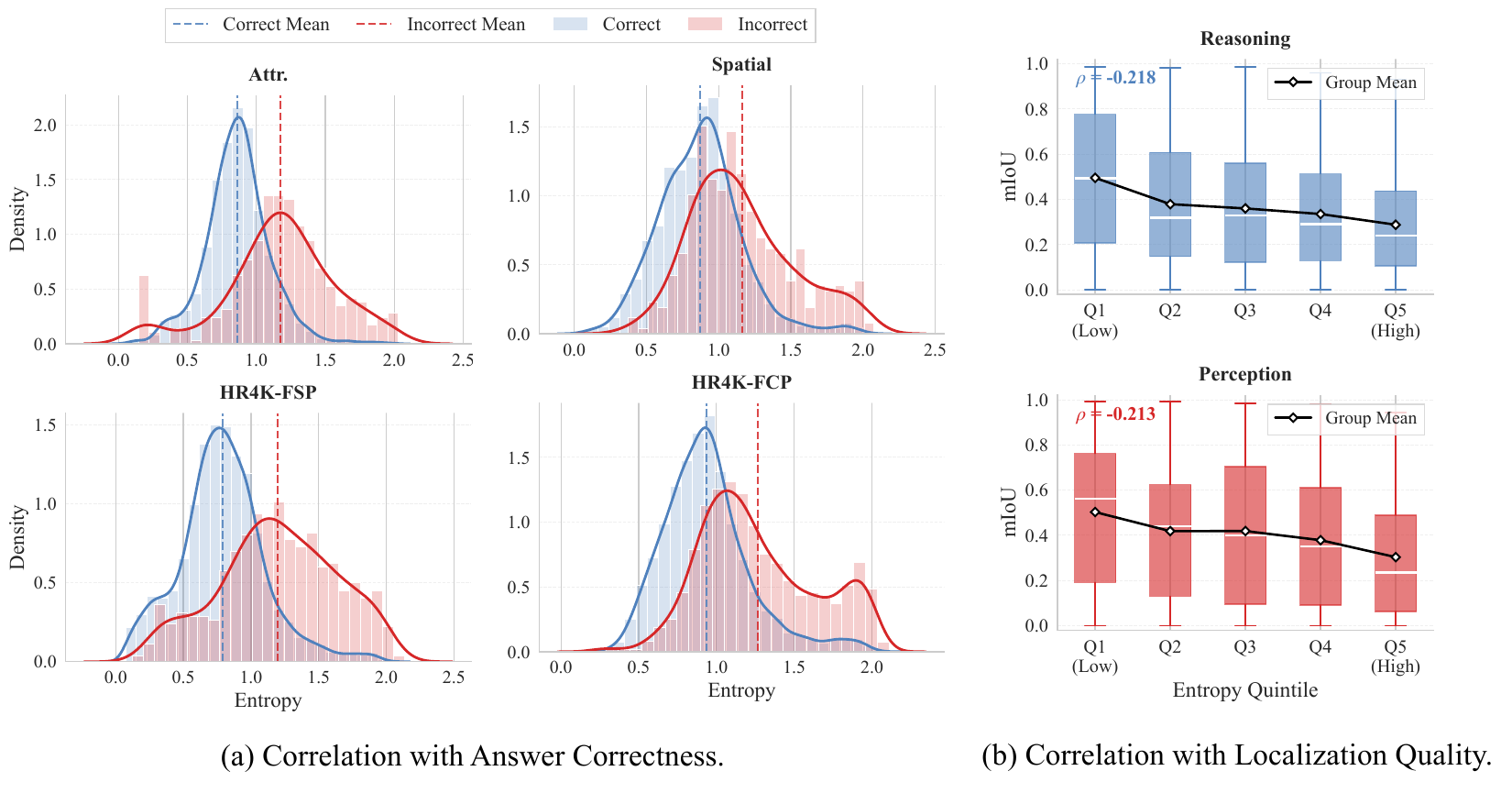}
    \caption{\textbf{Critical-token entropy tracks reliability.} Left:
    entropy-score distributions for correct and incorrect trajectories.
    Right: grounding mIoU by entropy quintile.}
    \label{fig:entropy_analysis}
  \end{minipage}\hfill
  \begin{minipage}[t]{0.34\linewidth}
    \centering
    \vspace{0.78in}
    \includegraphics[width=\linewidth]{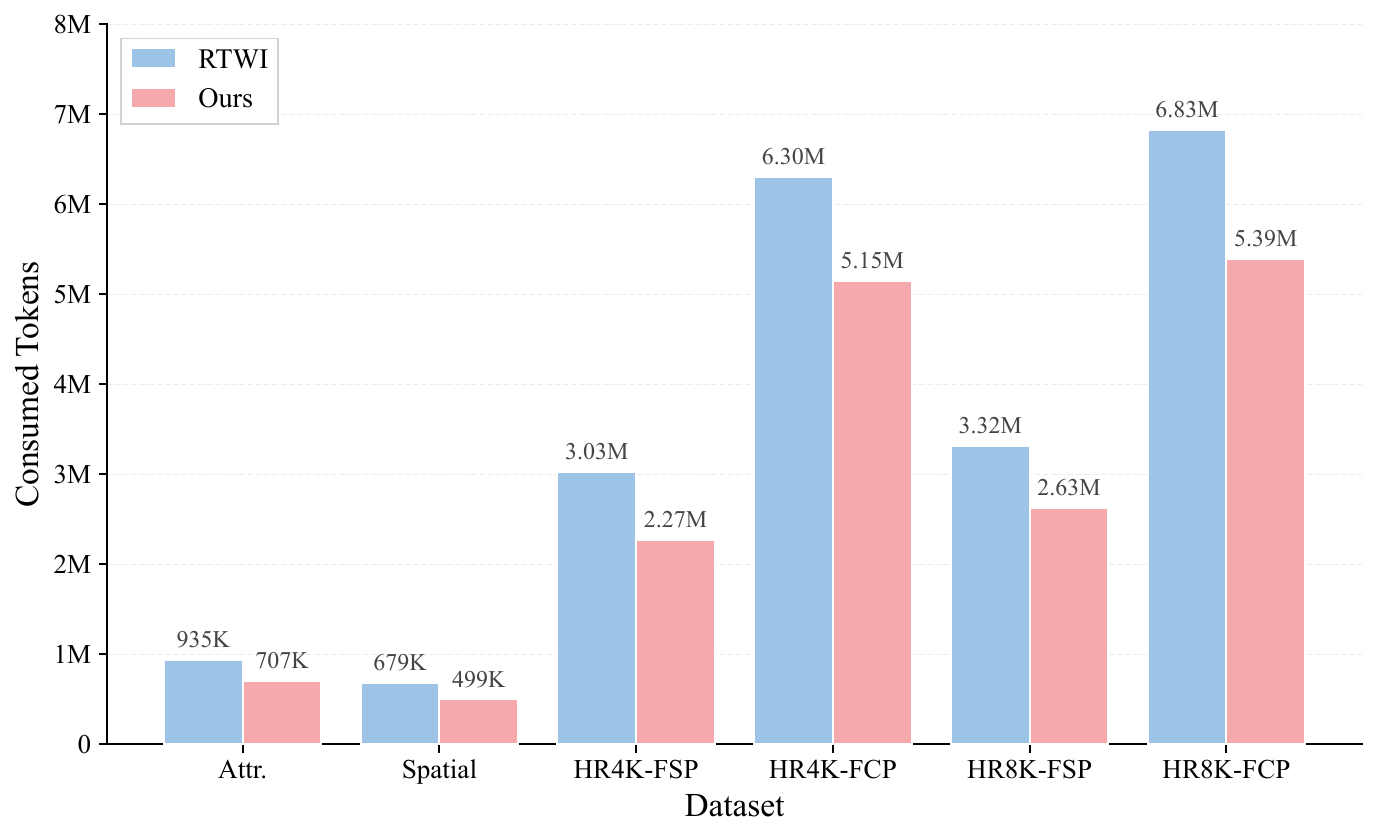}
    \caption{\textbf{Token consumption} of \method\ versus RTWI,
    averaged over five runs; ledger-update calls are included. Lower is
    better.}
    \label{fig:efficiency}
  \end{minipage}
\end{figure*}

\paragraph{Does the gate measure what we claim it measures?}
Figure~\ref{fig:entropy_analysis} tests the assumption behind
Eq.~\eqref{eq:score}: correct trajectories carry lower critical-token
entropy, and mIoU declines monotonically across entropy quintiles, so
the gate correlates not only with being right but with having looked in
the right place. Among five alternative reliability signals substituted
into the same pipeline (Appendix~\ref{sec:reliability_measures}),
critical-token entropy is the most consistent. The correlation is moderate rather than deterministic, which is why
\method\ never treats entropy as a verifier: a confident hallucination
is invisible to an intra-trajectory signal and is caught instead by the
cross-trajectory consensus rules of Section~\ref{sec:refinement}.

\paragraph{Do the gains come from better evidence or better voting?}
Because \method\ changes both what is observed and how observations are
combined, we measure grounding directly on TreeBench, which annotates
the required region, reporting round-by-round mIoU in
Appendix~\ref{sec:grounding_rounds}, computed per round over the fixed
number of traces retained in that round. The first round grounds less
accurately than the strongest baseline---the expected cost of diverse
exploration---but mIoU rises after every refinement round and overtakes
it on both splits. Since aggregation cannot alter which regions were
inspected, this attributes the gains to evidence acquisition rather
than to the readout rule.

\subsection{Scalability}
\label{sec:scalability}

Figure~\ref{fig:scaling} varies the three quantities that determine how
much computation \method\ consumes. Along \emph{backbone size}, all methods
improve and the ranking is unchanged, so perceptual scaling complements
rather than substitutes for parameter scaling; notably, the advantage does
not shrink as the backbone grows, indicating that a stronger model does not
resolve the grounding paradox on its own.
Appendix~\ref{sec:supp_other_backbones} finds the same ordering on two
backbones outside the Qwen3-VL family. Along \emph{width}, increasing $K$
helps every method, but \method\ improves faster than flat resampling:
width alone also enlarges the pool of unreliable observations, and its
benefit is realized only when the gate withholds them and the ledger
propagates the survivors; returns diminish at large $K$, as
Eq.~\eqref{eq:coverage} predicts. \emph{Depth} is the axis that
answer-level scaling does not possess: increasing $N$ spends compute on
making later hypotheses better informed rather than more numerous, and
because the fresh pool keeps sampling without prior commitments in every
round, depth does not come at the cost of coverage and performance keeps
improving.

\subsection{Efficiency}
\label{sec:efficiency}

\begin{figure*}[t]
  \begin{minipage}[t]{0.47\linewidth}
    \vspace{0pt}
    \centering
    \includegraphics[width=\linewidth]{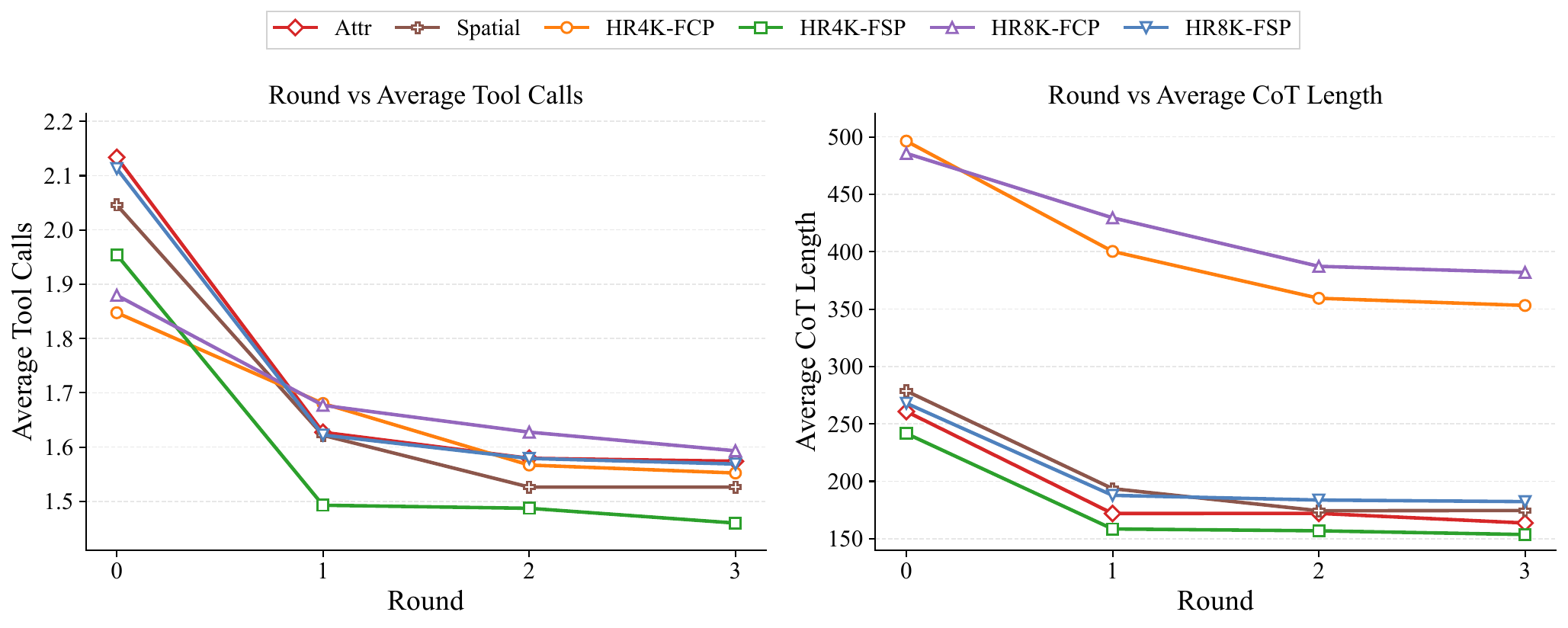}
    \caption{\textbf{Inference dynamics across rounds:} average crop
    invocations (left) and chain-of-thought length (right).}
    \label{fig:analysis}
  \end{minipage}\hfill
  \begin{minipage}[t]{0.51\linewidth}
    \vspace{0pt}
    \centering
    \includegraphics[width=\linewidth]{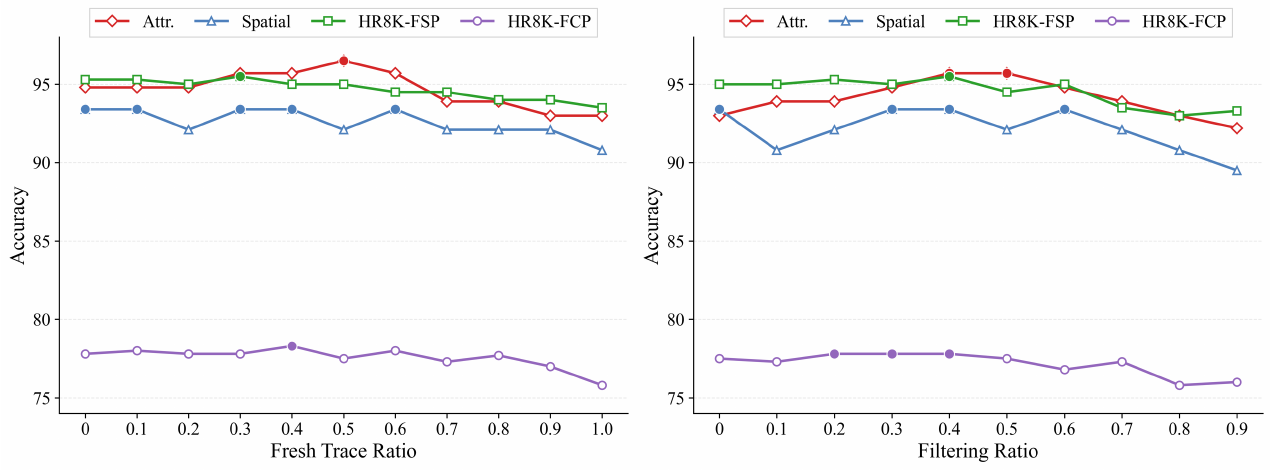}
    \caption{\textbf{Sensitivity to the fresh ratio $\alpha$ and the gate
    ratio $\rho$}, with other settings at their defaults.}
    \label{fig:hyperparameter}
  \end{minipage}
\end{figure*}

Generation scales as $\mathcal{O}(NKT_{\max})$, the same order as a
matched-budget multi-sample baseline plus $N-1$ greedy ledger updates, with
sequential depth $O(N)$; the full cost analysis is in
Appendix~\ref{app:algorithm}. Throughout, cost is measured in end-to-end
generated tokens, with the $N-1$ ledger-update calls and their prompts
included in \method's count; we note that wall-clock latency additionally
depends on the sequential depth $O(N)$, which single-round baselines do not
incur. Figure~\ref{fig:efficiency} compares
end-to-end generated tokens against RTWI, the strongest matched-budget
baseline: \method\ consumes fewer tokens on every subset even after
ledger-update calls are included, so accuracy and cost move in the same
direction. Figure~\ref{fig:analysis} shows that confirmed observations reduce
crop invocations and chain-of-thought length over rounds, redirecting compute
to contested regions.

\subsection{Hyperparameter Sensitivity}
\label{sec:hyper}

Figure~\ref{fig:hyperparameter} varies the two coupling ratios. Both curves
have an interior optimum, and the extremes fail for opposite reasons, as
the design predicts: small $\alpha$ removes the unconditioned channel
needed to overturn an erroneous ledger entry, while large $\alpha$ leaves
the ledger maintained but unread; weak gating admits unreliable
observations into both the ledger and the vote, while aggressive gating
withholds merely uncertain trajectories and narrows coverage.

\section{Conclusion}
\label{sec:conclusion}
We identified the \emph{grounding paradox} of thinking with images,
whose circularity makes early grounding errors self-reinforcing within
a trajectory and leaves answer-level test-time scaling
information-lossy. \method\ instead scales perception itself: it
samples diverse perceptual hypotheses, withholds unreliable ones, and
distills the rest into a correctable Evidence Ledger that steers later
rounds, outperforming matched-budget baselines across benchmarks and
backbones with fewer generated tokens. Grounding measurements attribute
the gains to better evidence rather than better voting; limitations are
discussed in Appendix~\ref{app:limitations}.


\bibliographystyle{assets/plainnat}
\bibliography{paper}

\newpage
\beginappendix

\section{Algorithm and Cost Analysis}
\label{app:algorithm}

Algorithm~\ref{alg:ttsp} states \method\ in full. Across rounds it maintains only
two states: the Evidence Ledger $\mathcal{E}_n$, consumed by the guided pool of
the next round, and the pool $\mathcal{F}$, which accumulates every trace that
survives gating and is consumed once at readout. Each round then advances the
three sub-problems of Section~\ref{sec:formulation} in one pass. For
\emph{coverage}, lines 4 to 6 split the budget and draw both pools from the same
policy under different conditioning contexts, with $\kappa_n=K$ in the first
round since $\mathcal{E}_0$ is empty. For \emph{selection}, lines 7 to 9 score
every trace from the log-probabilities already recorded during generation and
keep the highest-ranked $\lceil(1-\rho)K\rceil$. Ranking is performed within a
round, so no absolute threshold is ever needed, and a trace admitted early keeps
its vote. For \emph{utilization}, lines 10 to 13 refresh the ledger from the $m$
highest-scoring survivors and skip the update after the final round, because no
subsequent round would consume it.

\textsc{Sample} issues a batch of independent generations at temperature
$T_{\mathrm{gen}}$, inserting $\mathcal{E}$ into the system prompt when
non-empty, and \textsc{Score} evaluates Eq.~\eqref{eq:score} without any extra
forward pass. \textsc{UpdateLedger} applies the four rules of
Section~\ref{sec:refinement} through a single greedy-decoding call whose prompt
states them explicitly, and is deterministic because this step should summarize
collected evidence rather than add hypotheses. The procedure has three degenerate
settings, each of which is used as an ablation in
Section~\ref{sec:ablation_study}. Setting $\rho=0$ disables the gate, so every
sampled trace both votes and feeds the ledger. Setting $\alpha=1$ removes the
guided pool, so the ledger is still maintained but never read and depth reduces
to repeated independent sampling. Setting $\gamma\to\infty$ makes the readout an
unweighted majority vote.

\paragraph{Cost structure.}
A run costs $NK$ stochastic calls and $N-1$ greedy calls. With at most
$T_{\max}$ interaction turns per trajectory, generation scales as
$\mathcal{O}(NKT_{\max})$, the same order as a matched-budget multi-sample
baseline; the ledger updates are the only surplus. Scoring reuses the token
probabilities already recorded during generation, and ranking costs
$\mathcal{O}(K\log K)$ per round. Since the $K$ generations of a round are
independent given $\mathcal{E}_{n-1}$, they can be issued as a single batch and
the sequential depth of a run is $O(N)$ rather than $O(NK)$.

\begin{algorithm}[h]
\caption{\method: Test-Time Scaling over Perception}
\label{alg:ttsp}
\begin{algorithmic}[1]
\Require image $I$, question $Q$, options $\mathcal{O}$
\Statex \textbf{Hyperparameters:} rounds $N$, width $K$, fresh ratio
        $\alpha$, gate ratio $\rho$, vote temperature $\gamma$
\Ensure  answer $A^{*}$
\State $\mathcal{E}_0 \gets (\varnothing,\varnothing)$, \quad
       $\mathcal{F} \gets \varnothing$
       \Comment{empty ledger and retained pool}
\For{$n = 1$ \textbf{to} $N$}
    \Phase{Coverage}
    \State $\kappa_n \gets K$ \textbf{if} $n = 1$ \textbf{else}
           $\lceil \alpha K \rceil$
    \State $\mathcal{S}^{\text{fresh}} \gets$
           \Call{Sample}{$I,Q,\mathcal{O},\varnothing,\kappa_n$}
    \State $\mathcal{S}^{\text{guided}} \gets$
           \Call{Sample}{$I,Q,\mathcal{O},\mathcal{E}_{n-1},K-\kappa_n$}
    \Phase{Selection}
    \State $s(\tau) \gets$ \Call{Score}{$\tau$} \textbf{ for all }
           $\tau \in \mathcal{S}^{\text{fresh}} \cup
           \mathcal{S}^{\text{guided}}$
    \State $\mathcal{F}_n \gets$ the $\lceil (1-\rho)K \rceil$ traces of
           largest $s$
    \State $\mathcal{F} \gets \mathcal{F} \cup \mathcal{F}_n$
    \Phase{Utilization}
    \If{$n < N$}
        \State $\widetilde{\mathcal{F}}_n \gets$ the $m$ traces of
               $\mathcal{F}_n$ of largest $s$
        \State $\mathcal{E}_n \gets$ \Call{UpdateLedger}{$\mathcal{E}_{n-1},
               \widetilde{\mathcal{F}}_n,I,Q$}
    \EndIf
\EndFor
\Phase{Readout}
\State $W(a) \gets \sum_{\tau \in \mathcal{F} :\, a(\tau) \equiv a}
       \exp\!\bigl(s(\tau)/\gamma\bigr)$ \textbf{ for each answer class }
       $a$
\State \Return $A^{*} \gets \arg\max_{a} W(a)$
\end{algorithmic}
\end{algorithm}

\section{Experimental Setup Details}
\label{app:setup_details}

\paragraph{Benchmarks.}
The first regime targets high-resolution and fine-grained visual reasoning,
where the decisive evidence occupies a small fraction of the image: V$^{*}$
Bench \citep{wu2024vstar}, HR-Bench \citep{wang2025hrbench} at 4K and 8K
resolution, TreeBench \citep{wang2026TreeVGR}, and MME-RealWorld-Lite
\citep{zhang2024mme}. Several of these report perception and reasoning
subsets separately, which lets us ask whether better evidence acquisition
also benefits the inferences drawn from it. The second regime tests whether
the same procedure remains beneficial when perception is not the
bottleneck, using MMStar \citep{chen2024mmstar}, MMBench
\citep{liu2024mmbench}, MathVision \citep{wang2024mathvision}, LogicVista
\citep{xiao2024logicvista}, and MathVista \citep{lu2024mathvista}.

\paragraph{Baselines.}
The \emph{reference systems}, GPT-4o \citep{achiam2023gpt}, Thyme
\citep{zhang2025thyme}, DeepEyes \citep{zheng2025deepeyes}, and
Pixel-Reasoner \citep{wang2025pixel}, are trained or engineered for
tool-augmented visual reasoning but use different backbones and inference
budgets, so we report them to situate the difficulty of each suite rather
than as controlled comparisons. The \emph{test-time scaling methods},
self-consistency (SC) \citep{wang2022SC}, confidence-informed
self-consistency (CISC) \citep{taubenfeld2025CISC}, test-time recursive
thinking (TRT) \citep{zhuang2026test}, Self-Refine
\citep{madaan2023selfrefine}, Self-Certainty \citep{kang2025selfcertainty},
DeepConf \citep{fu2025deepconf}, and RTWI \citep{li2026RTWI}, are evaluated
on the same backbone and under the same trajectory budget as \method.

\paragraph{Budget protocol.}
All test-time scaling baselines are matched on the number of sampled
trajectories, so single-round methods draw $NK$ traces in one shot while
multi-round methods receive the same total. \method\ additionally issues
$N-1$ greedy ledger updates outside this matched count; because this is a
small but real surplus, the end-to-end generated-token comparison of
Section~\ref{sec:efficiency} includes the ledger overhead in the
accounting. Unless stated otherwise we use $N=4$ rounds with $K=8$ traces
per round, a fresh ratio $\alpha=0.4$, and a gate ratio $\rho=0.4$. Traces
are sampled at temperature $1.0$, whereas every ledger update uses greedy
decoding.

\begin{table}[H]
\centering
\caption{Category-level results on TreeBench. Perception covers attributes
(Attr.), material (Mat.), physical state (Phys.), object retrieval (Ret.), and
OCR. Reasoning covers perspective transformation (Persp.), ordering (Ord.),
contact and occlusion (Cont.), spatial containment (Cntm.), and comparison
(Comp.).}
\label{tab:tree_bench}
\small
\setlength{\tabcolsep}{7pt}
\resizebox{\linewidth}{!}{%
\begin{tabular}{l cccccc cccccc}
\toprule
& \multicolumn{6}{c}{Perception} & \multicolumn{6}{c}{Reasoning} \\
\cmidrule(lr){2-7} \cmidrule(lr){8-13}
Method & Attr. & Mat. & Phys. & Ret. & OCR & Ovr. & Persp. & Ord. & Cont. & Cntm. & Comp. & Ovr.\\
\midrule
GPT-4o & 51.7 & 61.5 & 65.2 & 43.8 & 69.1 & 61.7 & 18.8 & 38.6 & 48.8 & 72.4 & 43.2 & 38.3 \\
DeepEyes & 62.1 & 53.8 & 65.2 & 68.8 & 51.5 & 57.7 & 11.8 & 24.6 & 36.6 & 51.7 & 47.7 & 29.3 \\
Pixel-Reasoner & 58.6 & 61.5 & 65.2 & 50.0 & 48.5 & 54.3 & 14.1 & 31.6 & 39.0 & 44.8 & 40.9 & 30.1\\
\midrule
\rowcolor{myblue}
\multicolumn{13}{c}{\textit{Qwen3-VL-4B-Instruct}} \\
\midrule
Base & 44.8 & 61.5 & 52.2 & 62.5 & 69.1 & 60.4 & 15.3 & 29.8 & 48.8 & 75.9 & 47.7 & 36.3 \\
SC & 51.7 & 61.5 & 56.5 & 62.5 & 73.5 & 64.4 & 16.5 & \textbf{31.6} & 51.2 & 75.9 & \textbf{50.0} & 37.9 \\
CISC & 55.2 & 61.5 & 52.2 & 62.5 & 72.1 & 63.8 & 16.5 & \textbf{31.6} & 51.2 & 75.9 & 47.7 & 37.5 \\
TRT & 51.7 & 61.5 & 56.5 & 62.5 & 72.1 & 63.8 & 16.5 & \textbf{31.6} & 51.2 & 75.9 & 47.7 & 37.5 \\
Self-Ref. & 51.7 & 61.5 & 56.5 & 62.5 & 73.5 & 64.4 & 16.5 & \textbf{31.6} & 51.2 & 75.9 & \textbf{50.0} & 37.9 \\
Self-Cer. & 51.7 & 61.5 & 52.2 & 62.5 & 73.5 & 63.7 & 15.3 & \textbf{31.6} & 48.8 & 75.9 & \textbf{50.0} & 37.5 \\
DeepConf & 55.2 & 61.5 & 56.5 & \textbf{68.8} & 73.5 & 65.8 & 16.5 & \textbf{31.6} & 51.2 & \textbf{79.3} & \textbf{50.0} & 37.9 \\
RTWI & 58.6 & 61.5 & 56.5 & 62.5 & 75.0 & 66.4 & 17.6 & \textbf{31.6} & \textbf{53.6} & \textbf{79.3} & \textbf{50.0} & 39.0 \\
\rowcolor{mygrey}
\textbf{\method} & \textbf{62.1} & \textbf{69.2} & \textbf{60.9} & \textbf{68.8} & \textbf{77.9} & \textbf{70.5} & \textbf{18.8} & \textbf{31.6} & \textbf{53.6} & \textbf{79.3} & \textbf{50.0} & \textbf{39.4}\\
\midrule
\rowcolor{myblue}
\multicolumn{13}{c}{\textit{Qwen3-VL-8B-Instruct}} \\
\midrule
Base & 51.7 & 61.5 & 60.9 & 68.8 & 76.5 & 67.1 & 15.3 & 26.3 & 46.3 & 72.4 & 45.5 & 34.4 \\
SC & 55.2 & 61.5 & \textbf{65.2} & 75.0 & 77.9 & 69.8 & 16.5 & 29.8 & 48.8 & 75.9 & \textbf{50.0} & 37.1 \\
CISC & 55.2 & 61.5 & 60.9 & 68.8 & 76.5 & 67.8 & 15.3 & 29.8 & \textbf{51.2} & 75.9 & 47.7 & 36.7 \\
TRT & 55.2 & 61.5 & 60.9 & 75.0 & 77.9 & 69.1 & 16.5 & 29.8 & 48.8 & 75.9 & \textbf{50.0} & 37.1 \\
Self-Ref. & 55.2 & 61.5 & 60.9 & 75.0 & 77.9 & 69.1 & 16.5 & 31.6 & 48.8 & 72.4 & 47.7 & 36.7\\
Self-Cer. & 55.2 & 61.5 & 60.9 & 68.8 & 76.5 & 67.8 & 15.3 & 31.6 & 48.8 & 75.9 & 47.7 & 36.7 \\
DeepConf & 58.6 & 61.5 & \textbf{65.2} & 75.0 & 77.9 & 70.4 & 16.5 & 33.3 & \textbf{51.2} & 75.9 & 47.7 & 37.9 \\
RTWI & 58.6 & \textbf{69.2} & \textbf{65.2} & 75.0 & 77.9 & 71.1 & 16.5 & 33.3 & 48.8 & 75.9 & \textbf{50.0} & 37.9 \\
\rowcolor{mygrey}
\textbf{\method} & \textbf{62.1} & \textbf{69.2} & \textbf{65.2} & \textbf{81.3} & \textbf{79.4} & \textbf{73.2} & \textbf{17.6} & \textbf{35.1} & \textbf{51.2} & \textbf{79.3} & \textbf{50.0} & \textbf{39.4}\\
\bottomrule
\end{tabular}
}
\end{table}

\section{Additional Experimental Results}

\subsection{Category-Level Results on TreeBench}
\label{sec:tree_detailed}

Table~\ref{tab:tree_bench} expands the three TreeBench columns of
Table~\ref{tab:main_result}. The Perception split contains $149$ questions
covering attributes, material, physical state, object retrieval, and OCR, and the
Reasoning split contains $256$ questions covering perspective transformation,
ordering, contact and occlusion, spatial containment, and comparison. The
main-table Overall score micro-averages the two splits over all $405$ questions,
so the two split scores and the overall score are mutually consistent by
construction.

Two patterns are worth noting. \method\ attains the best score in every
perception category, and the categories in which it separates most clearly from
the strongest baseline are those whose answer depends on reading a small or
partially occluded region rather than on identifying a salient object. On the
reasoning side it matches or exceeds the best baseline in every category, and the
categories that improve are the ones whose premises are supplied by perception,
such as ordering and contact relations, rather than those that are dominated by
a single relational judgment. This ordering is what the intended causal chain
predicts, in which better localized evidence first improves perception and then
becomes available to the reasoning that consumes it.

\subsection{Category-Level Results on MME-RealWorld-Lite}
\label{sec:mme_detailed}

Table~\ref{tab:mme_detailed} expands the three MME-RealWorld-Lite columns of
Table~\ref{tab:main_result}. The Perception split contains $1{,}169$ examples
drawn from OCR ($250$), remote sensing ($150$), diagram and table understanding
($100$), monitoring ($319$), and autonomous driving ($350$). The Reasoning split
contains $750$ examples drawn from OCR ($100$), diagram and table understanding
($100$), monitoring ($150$), and autonomous driving ($400$). The main-table
Overall score micro-averages the two splits over all $1{,}919$ examples.

The breakdown separates two regimes that the aggregate score conflates. On
categories where the target is legible in the global view, such as OCR and
diagram understanding, all matched-budget methods are close to one another and
close to saturation, and \method\ offers little room to improve. The categories
in which it separates most clearly are remote sensing, monitoring, and autonomous
driving, all of which present a wide field of view in which the queried object
covers a small share of the frame. This is the regime the grounding paradox
describes, and it is also where the largest share of the aggregate improvement
originates. The same asymmetry appears on the Reasoning split, which indicates
that the benefit is not confined to recognition but carries into the inferences
that consume the recognized evidence.

\begin{table*}[htbp]
\centering
\caption{Category-level results on MME-RealWorld-Lite. RS, DT, MO, and AD denote
remote sensing, diagram and table understanding, monitoring, and autonomous
driving, respectively.}
\label{tab:mme_detailed}
\small
\setlength{\tabcolsep}{8.5pt}
\begin{tabular}{l cccccc ccccc}
\toprule
& \multicolumn{6}{c}{Perception} & \multicolumn{5}{c}{Reasoning} \\
\cmidrule(lr){2-7} \cmidrule(lr){8-12}
Method & OCR & RS & DT & MO & AD & Ovr.
& OCR & DT & MO & AD & Ovr. \\
\midrule
Pixel-Reasoner & 89.6 & 52.0 & 86.0 & 38.9 & 30.9 & 53.1
& 71.0 & 72.0 & 46.0 & 32.5 & 45.6 \\
DeepEyes & 90.0 & 52.7 & 89.0 & 43.3 & 33.4 & 55.4
& 76.0 & 69.0 & 44.0 & 35.0 & 46.8 \\
\midrule
\rowcolor{myblue}
\multicolumn{12}{c}{\textit{Qwen3-VL-4B-Instruct}} \\
\midrule
Base & 89.6 & 50.7 & 85.0 & 37.3 & 34.3 & 53.4
& 73.0 & 75.0 & 42.7 & 31.0 & 44.8 \\
SC & 91.2 & 53.3 & 89.0 & 38.6 & 36.9 & 55.5
& 78.0 & 78.0 & 43.3 & 32.8 & 47.0 \\
CISC & 91.2 & 53.3 & 89.0 & 38.6 & 37.4 & 55.7
& 78.0 & 78.0 & 42.7 & 32.3 & 46.6 \\
TRT & 90.8 & 53.3 & 89.0 & 38.6 & 36.9 & 55.5
& 77.0 & 79.0 & 43.3 & 32.8 & 47.0 \\
Self-Ref. & 91.2 & 53.3 & 89.0 & 38.2 & 36.9 & 55.4
& 78.0 & 78.0 & 42.7 & 32.3 & 46.6 \\
Self-Cer. & 90.8 & 54.0 & 89.0 & 38.6 & 36.9 & 55.5
& 78.0 & 78.0 & 42.7 & 32.8 & 46.8 \\
DeepConf & 91.6 & 53.3 & 89.0 & 38.9 & 37.4 & 55.9
& 78.0 & 79.0 & 43.3 & 32.8 & 47.1 \\
RTWI & 92.4 & 54.7 & 89.0 & 39.2 & 38.3 & 56.6
& 79.0 & 80.0 & 43.3 & 32.8 & 47.2 \\
\rowcolor{mygrey}
\textbf{\method} & \textbf{93.2} & \textbf{56.0} & \textbf{91.0}
& \textbf{39.8} & \textbf{40.0} & \textbf{57.7}
& \textbf{81.0} & \textbf{83.0} & \textbf{44.0}
& \textbf{34.3} & \textbf{49.0} \\
\midrule
\rowcolor{myblue}
\multicolumn{12}{c}{\textit{Qwen3-VL-8B-Instruct}} \\
\midrule
Base & 89.2 & 54.7 & 86.0 & 34.8 & 33.1 & 52.9
& 74.0 & 80.0 & 34.0 & 35.3 & 46.2 \\
SC & 90.0 & 58.0 & 89.0 & 39.2 & 38.6 & 56.6
& 81.0 & 83.0 & 40.7 & 36.0 & 49.2 \\
CISC & 90.0 & 60.0 & 89.0 & 39.2 & 38.6 & 56.8
& 80.0 & 83.0 & 40.7 & 36.0 & 49.1 \\
TRT & 90.0 & 60.0 & 89.0 & 39.2 & 38.6 & 56.8
& 81.0 & 83.0 & 40.7 & 36.0 & 49.2 \\
Self-Ref. & 90.4 & 60.0 & 89.0 & 38.9 & 38.3 & 56.7
& 82.0 & 83.0 & 40.0 & 36.3 & 49.4 \\
Self-Cer. & 90.0 & 60.0 & 92.0 & 39.2 & 38.6 & 57.1
& 81.0 & 83.0 & 40.7 & 36.3 & 49.4 \\
DeepConf & 90.0 & 61.3 & 90.0 & 39.5 & 38.9 & 57.2
& 82.0 & 83.0 & 41.3 & 36.5 & 49.7 \\
RTWI & 90.4 & 62.7 & 92.0 & 39.8 & 38.9 & 57.8
& 83.0 & 84.0 & 42.0 & 36.5 & 50.1 \\
\rowcolor{mygrey}
\textbf{\method} & \textbf{90.8} & \textbf{65.3} & \textbf{94.0}
& \textbf{41.1} & \textbf{40.0} & \textbf{59.0}
& \textbf{84.0} & \textbf{88.0} & \textbf{45.3}
& \textbf{38.8} & \textbf{52.7} \\
\bottomrule
\end{tabular}
\end{table*}

\subsection{Grounding Quality Across Rounds}
\label{sec:grounding_rounds}

Accuracy alone cannot tell whether \method\ wins because it observes better
evidence or because it combines a fixed set of observations better.
Table~\ref{tab:grounding_rounds} therefore measures the inspected regions
directly on TreeBench, which annotates the region required by each question, and
reports the mean IoU between the annotated region and the regions a method
actually inspects. The measurement is taken after each round of \method\ and is
compared with the strongest matched-budget baseline. The mIoU at round $n$ is
computed over the $\lceil(1-\rho)K\rceil$ traces retained in round $n$ alone,
not over the cumulative pool; since gating retains the same number of traces in
every round, the number of candidate regions entering each measurement is
identical across rounds, and the round-wise increase cannot be an artifact of an
accumulating candidate pool.

Three observations follow. In its first round \method\ grounds less accurately
than the baseline, which is the expected cost of a deliberately diverse set of
hypotheses rather than a single confident guess. The mIoU then increases after
every refinement round on both splits, and it overtakes the baseline before the
budget is exhausted. Since the readout rule cannot change which regions were
inspected, the monotone increase is attributable to the ledger rather than to
aggregation, and it gives a concrete reading of what refinement does, namely
converting an initially imprecise set of perceptual hypotheses into progressively
better localized evidence. The reasoning split remains above the perception split
throughout, which reflects that its questions more often name a large or
relationally defined region and are therefore easier to cover.

\begin{table}[htbp]
\centering
\caption{Grounding mIoU on TreeBench after each round of \method, compared with
the strongest matched-budget baseline. R1 to R4 denote the refinement rounds;
each round is measured over the same number of retained traces.}
\label{tab:grounding_rounds}
\small
\setlength{\tabcolsep}{5pt}
\begin{tabular}{lccccc}
\toprule
Split & RTWI & R1 & R2 & R3 & R4 \\
\midrule
Perception & 40.65 & 37.83 & 40.38 & 42.03 & \textbf{45.29} \\
Reasoning  & 45.82 & 42.79 & 44.51 & 47.04 & \textbf{49.22} \\
\bottomrule
\end{tabular}
\end{table}

\subsection{Alternative Reliability Signals}
\label{sec:reliability_measures}

Section~\ref{sec:analysis} shows that critical-token entropy correlates with both
answer correctness and grounding quality, but it does not by itself establish
that this particular statistic is the right choice.
Table~\ref{tab:reliability_measures} therefore replaces it inside the same
pipeline, holding the backbone, the trajectory budget, the gate ratio, and the
readout rule fixed, and substituting five alternative signals: the probability of
the final answer, the margin between the top two answer probabilities, the mean
token probability, the length-normalized sequence probability, and cross-trace
agreement.

Critical-token entropy dominates across the board. It matches or exceeds every
alternative on all six metrics and attains the highest average by a clear margin, whereas each competing signal is at best competitive on a
few subsets and strictly weaker elsewhere. This gap is exactly what the design of
the signal predicts: three of the alternatives summarize confidence over the
whole sequence and are therefore diluted by routine tokens, one reads confidence
only at the answer position and ignores how the evidence was obtained, and
cross-trace agreement is an inter-trace signal that is blind to whether an
individual trace committed stably. Critical-token entropy instead reads
confidence exactly at the positions where a trace commits to a region and to an
answer, which is why it is uniformly the most reliable verifier in this study.
We therefore adopt it as the default signal inside \method, and further pair it
with cross-trace consensus in the ledger so that its per-trace reliability
estimate is reinforced by inter-trace agreement.

\begin{table*}[t]
\centering
\caption{Accuracy when alternative trace-reliability signals replace
critical-token entropy inside \method, with all other settings fixed. Results use
Qwen3-VL-4B-Instruct. Avg.\ is the mean of the six reported metrics.}
\label{tab:reliability_measures}
\small
\setlength{\tabcolsep}{7pt}
\begin{tabular}{lccccccc}
\toprule
& \multicolumn{2}{c}{V$^{*}$ Bench}
& \multicolumn{2}{c}{HR-Bench-4K}
& \multicolumn{2}{c}{HR-Bench-8K} & \\
\cmidrule(lr){2-3} \cmidrule(lr){4-5} \cmidrule(lr){6-7}
Signal & Attr. & Spat. & FSP & FCP & FSP & FCP & Avg. \\
\midrule
Answer probability & 93.9 & 92.1 & 96.0 & 75.3 & 93.0 & 74.5 & 87.5 \\
Probability margin & 93.0 & \textbf{93.4} & 96.5 & 75.8 & 93.5 & 74.5 & 87.8 \\
Mean token probability & \textbf{94.8} & 92.1 & 96.0 & 76.3 & 93.0 & 74.0 & 87.7 \\
Length-normalized probability & 93.9 & 92.1 & 96.5 & 76.3 & 93.5 & 74.5 & 87.8 \\
Cross-trace agreement & 93.9 & 92.1 & 96.0 & \textbf{76.5} & 93.5 & \textbf{74.8} & 87.8 \\
\midrule
\rowcolor{mygrey}
\textbf{Critical-token entropy} & \textbf{94.8} & \textbf{93.4} & \textbf{96.8}
& \textbf{76.5} & \textbf{94.0} & \textbf{74.8} & \textbf{88.4} \\
\bottomrule
\end{tabular}
\end{table*}

\subsection{Generalization to Additional Backbones}
\label{sec:supp_other_backbones}

The main experiments use two instruction-tuned backbones from a single family, so
Table~\ref{tab:supp_results} repeats the comparison on two further backbones. The
first is DeepEyes-7B, built on Qwen2.5-VL-7B-Instruct, which is of particular
interest because it was explicitly trained for tool-augmented perception and
therefore already allocates crops competently on its own. The second is
Qwen3.5-9B, which places the comparison at a different scale and in a different
model generation. Absolute scores are not comparable with the main tables, since
the backbones, the crop interface, and the prompt formats differ, and the
DeepEyes row of Table~\ref{tab:main_result} refers to the released system under
its own inference setup rather than to the reproduction used here.

\method\ ranks first on all four benchmarks with both backbones, and the ordering
among the baselines is broadly preserved. The DeepEyes-7B result is the more
informative of the two. A backbone trained to invoke crops does not remove the
grounding paradox, because it still has to select a first region from the global
view under incomplete evidence, and the improvement obtained by scaling
perception rather than answers is therefore not an artifact of using a backbone
that lacks tool-use ability. Taken together with the model-size panel of
Figure~\ref{fig:scaling}, this indicates that the procedure is not tied to the
model family used in the main experiments.

\begin{table*}[t]
\centering
\caption{Generalization of \method\ to additional multimodal backbones on
high-resolution benchmarks. Absolute scores are not comparable with
Table~\ref{tab:main_result}, since the backbones and inference setups differ.}
\label{tab:supp_results}
\small
\setlength{\tabcolsep}{7pt}
\begin{tabular}{l ccc ccc ccc ccc}
\toprule
& \multicolumn{3}{c}{V$^{*}$ Bench} & \multicolumn{3}{c}{HR-Bench-4K} & \multicolumn{3}{c}{HR-Bench-8K} & \multicolumn{3}{c}{TreeBench}\\
\cmidrule(lr){2-4} \cmidrule(lr){5-7} \cmidrule(lr){8-10} \cmidrule(lr){11-13}
Method & Attr. & Spat. & Ovr. & FSP & FCP & Ovr. & FSP & FCP & Ovr. & Perc. & Reas. & Ovr.\\
\midrule
\rowcolor{myblue}
\multicolumn{13}{c}{\textit{DeepEyes-7B}} \\
\midrule
Base & 87.8 & 81.0 & 85.1 & 90.0 & 58.5 & 74.2 & 86.5 & 57.3 & 71.9 & 57.0 & 35.9 & 43.7\\
SC & 89.6 & 82.3 & 86.7 & 91.3 & 61.0 & 76.2  & 88.0 & 61.0 & 74.5 & 61.1 & 37.1 & 45.9 \\
CISC & 90.4 & 82.3 & 87.2 & 91.3 & 59.0 & 75.2 & 88.0 & 61.8 & 74.9 & 61.1 & 37.5 & 46.2\\
TRT & 89.6 & 81.0 & 86.2 &  90.0 & 61.8 & 75.9 & 87.8 & 59.0 & 73.4 & 60.4 & 37.1 & 45.7\\
Self-Ref. & 90.4 & 82.3 & 87.2 & 91.5 & 61.5 & 76.5 & 88.3 & 61.5 & 74.9 & 61.1  & 37.9 & 46.4\\
Self-Cer. & 89.6 & 82.3 & 86.7 & 91.5 & 59.5 & 75.5 & 87.8 & 61.0 & 74.4 & 61.7 & 37.5 & 46.4\\
DeepConf & 90.4 & 82.3 & 87.2 & 92.0 & 61.8 & 76.9 & 88.5 & 61.5 & 75.0 & 62.4 & 37.5 & 46.7\\
RTWI & 90.4 & 82.3 & 87.2 & 92.3 & 62.3 & 77.3 & 88.8 & 61.8 & 75.3 & 62.4 & 37.9 & 46.9\\
\rowcolor{mygrey}
\textbf{\method} & \textbf{92.2} & \textbf{83.5} & \textbf{88.7} & \textbf{92.8} & \textbf{63.0} &  \textbf{77.9} & \textbf{89.8} & \textbf{62.3} & \textbf{76.0} & \textbf{64.4} & \textbf{38.7} & \textbf{48.2}\\
\midrule
\rowcolor{myblue}
\multicolumn{13}{c}{\textit{Qwen3.5-9B}} \\
\midrule
Base & 82.6 & 86.8 & 84.3 & 85.8 & 77.0 & 81.4 & 79.0 & 75.5 & 77.2 & 67.1 & 36.3 & 47.7 \\
SC & 86.1 & 90.8 & 88.0 & 90.3 & 78.5 & 84.4 & 89.8 & 76.5 & 83.2 & 73.8 & 39.4 & 52.1 \\
CISC & 86.1 & 90.8 & 88.0 & 91.0 & 78.5 & 84.8 & 89.8 & 76.5 & 83.2 & 73.8 & 39.1 & 51.9 \\
TRT & 86.1 & 90.8 & 88.0 & 90.8 & 78.5 & 84.7 & 88.3 & 76.8 & 82.5 & 73.5 & 39.4 & 51.9 \\
Self-Ref. & 87.8 & 92.1 & 89.5 & 91.5 & 78.5 & 85.0 & 87.5 & 76.3 & 81.9 & 74.8 & 39.1 & 52.2 \\
Self-Cer. & 86.1 & 92.1 & 88.5 & 90.3 & 78.3 & 84.3 & 89.8 & 76.8 & 83.3 & 75.2 & 39.4 & 52.6 \\
DeepConf & 87.0 & 92.1 & 89.0  & 91.5 & 78.5 & 85.0 & 90.8 & 77.0 & 83.9  & 75.2 & 40.2 & 53.1 \\
RTWI & 87.0 & 93.4 & 89.5 & 91.8 & 79.0 & 85.4 & 91.0 & 77.3 & 84.2 & 75.8 & 39.8 & 53.0 \\
\rowcolor{mygrey}
\textbf{\method} & \textbf{88.7} & \textbf{96.1} & \textbf{91.6} & \textbf{93.8} & \textbf{79.8} & \textbf{86.8} & \textbf{91.5} & \textbf{78.5} & \textbf{85.0} & \textbf{78.6} & \textbf{42.6} & \textbf{55.8} \\
\bottomrule
\end{tabular}
\end{table*}

\subsection{Additional Results on VisualProbe}
\label{sec:visualprobe}

VisualProbe \citep{lai2025mini} complements the main evaluation by
stratifying fine-grained perception into Easy, Medium, and Hard splits according
to the difficulty of locating the queried content, which makes it a direct test
of whether the benefit of scaling perception grows with the severity of the
grounding problem. Table~\ref{tab:visualprobe} shows that \method\ is best on
every split, and that the ordering among the baselines is less stable here than
on the aggregate benchmarks, with several answer-level methods falling below
plain self-consistency. This is consistent with the picture in
Section~\ref{sec:results}: when a single trajectory rarely inspects the required
region, reweighting or revising answers over a fixed set of trajectories has
little to work with. The margin over the strongest baseline is widest on the Hard
split, which is the intended regime, and narrowest on the Easy split, where a
single global view is often already sufficient.

\begin{table}[t]
\centering
\caption{Results on VisualProbe with Qwen3-VL-8B-Instruct.}
\label{tab:visualprobe}
\small
\setlength{\tabcolsep}{7pt}
\begin{tabular}{lcccc}
\toprule
Method & Easy & Medium & Hard & Overall \\
\midrule
Base      & 56.0 & 35.4 & 34.9 & 40.9 \\
SC        & 64.0 & 40.2 & 42.5 & 47.2 \\
CISC      & 59.7 & 38.6 & 39.6 & 44.6 \\
TRT       & 59.7 & 39.6 & 40.6 & 45.3 \\
Self-Ref. & 61.0 & 39.6 & 38.7 & 45.3 \\
Self-Cer. & 59.0 & 39.4 & 37.7 & 44.4 \\
DeepConf  & 61.9 & 40.6 & 43.4 & 47.0 \\
RTWI      & 65.4 & 40.6 & 42.5 & 47.8 \\
\midrule
\rowcolor{mygrey}
\textbf{\method} & \textbf{67.4} & \textbf{41.4} & \textbf{47.2} & \textbf{49.7} \\
\bottomrule
\end{tabular}
\end{table}

\section{Implementation Details}

\subsection{Decoding and Infrastructure}
\label{sec:generation_hyperparameters}

Trace generation and ledger construction use different decoding regimes by
design. All traces are sampled with temperature $1.0$, top-$p$ of $1.0$, and
top-$k$ disabled, which supplies the diversity that coverage requires, with a
maximum sequence length of $51{,}200$ tokens. Every ledger update instead uses
greedy decoding, because that step should summarize the evidence already
collected rather than introduce new hypotheses. The reliability score of
Eq.~\eqref{eq:score} is computed from the $k_{\mathrm{top}}$ largest
log-probabilities recorded at each step during generation, averaged over the
$k'$ positions of largest truncated entropy. The same settings are used for
every backbone and every benchmark, and no per-dataset tuning is performed;
Table~\ref{tab:hyperparams} summarizes the shared defaults. All experiments run
on $8\times$H20 GPUs, and trajectories within a round are issued as a single
batch, which is possible because they are conditionally independent given the
ledger.

\begin{table}[t]
\centering
\caption{Default settings of \method, shared across all backbones and
benchmarks.}
\label{tab:hyperparams}
\small
\setlength{\tabcolsep}{7pt}
\begin{tabular}{ll}
\toprule
Setting & Value \\
\midrule
Perception depth $N$ & $4$ \\
Perception width $K$ & $8$ \\
Fresh ratio $\alpha$ & $0.4$ \\
Gate ratio $\rho$ & $0.4$ \\
Entropy truncation $k_{\mathrm{top}}$ & $256$ \\ 
Critical positions $k'$ & $128$ \\ 
Vote temperature $\gamma$ & $1.0$ \\ 
Ledger context size $m$ & $4$ \\ 
Confirmed-tier cap $v_{\max}$ & $8$ \\ 
Conflict-tier cap $u_{\max}$ & $4$ \\ 
Sampling temperature $T_{\mathrm{gen}}$ & $1.0$ \\
Top-$p$ / top-$k$ & $1.0$ / disabled \\
Maximum sequence length & $51{,}200$ tokens \\
Ledger-update decoding & greedy \\
\bottomrule
\end{tabular}
\end{table}

\subsection{Prompts}
\label{sec:prompts}

\begin{figure*}[t]
  \centering
  \includegraphics[width=0.9\linewidth]{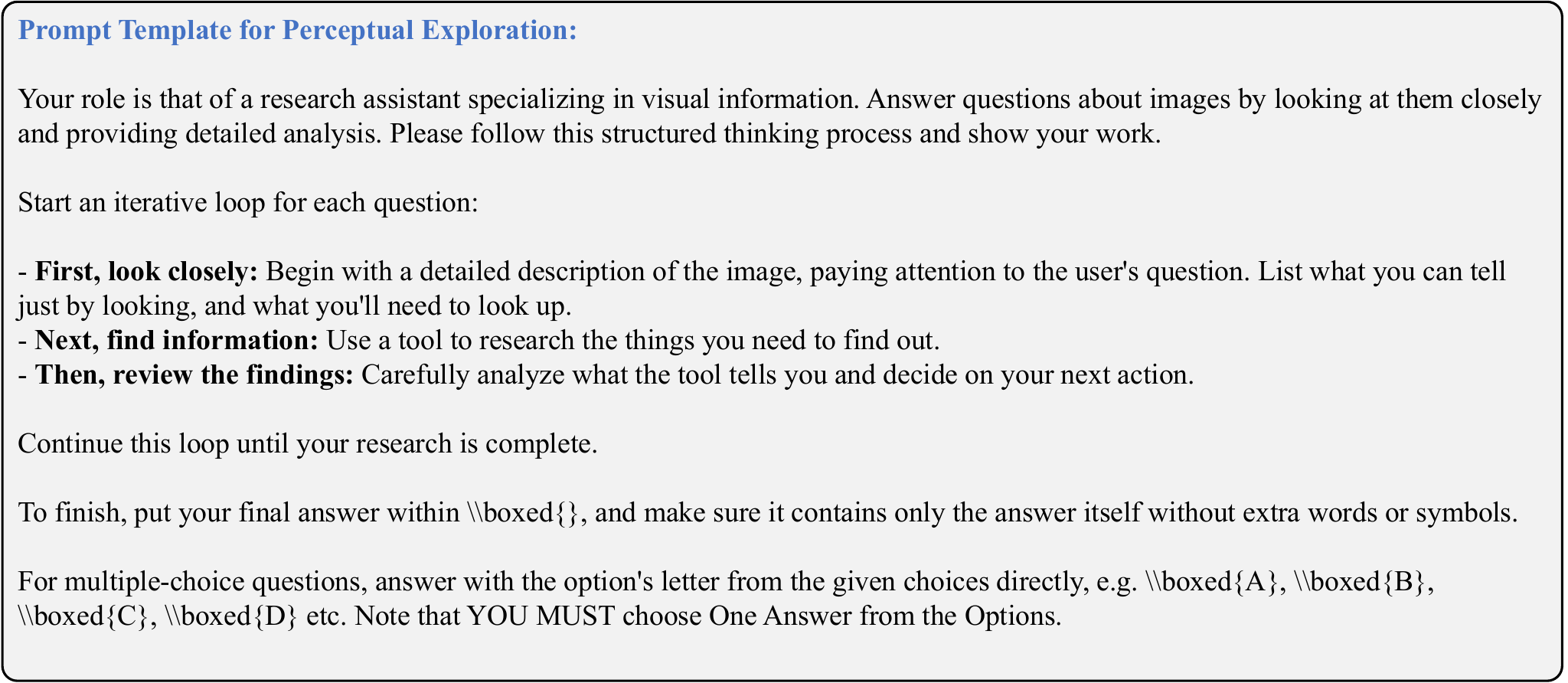}
  \caption{Prompt template for perceptual exploration.}
  \label{fig:perceptual_exploration}
\end{figure*}

\begin{figure*}[t]
  \centering
  \includegraphics[width=0.9\linewidth]{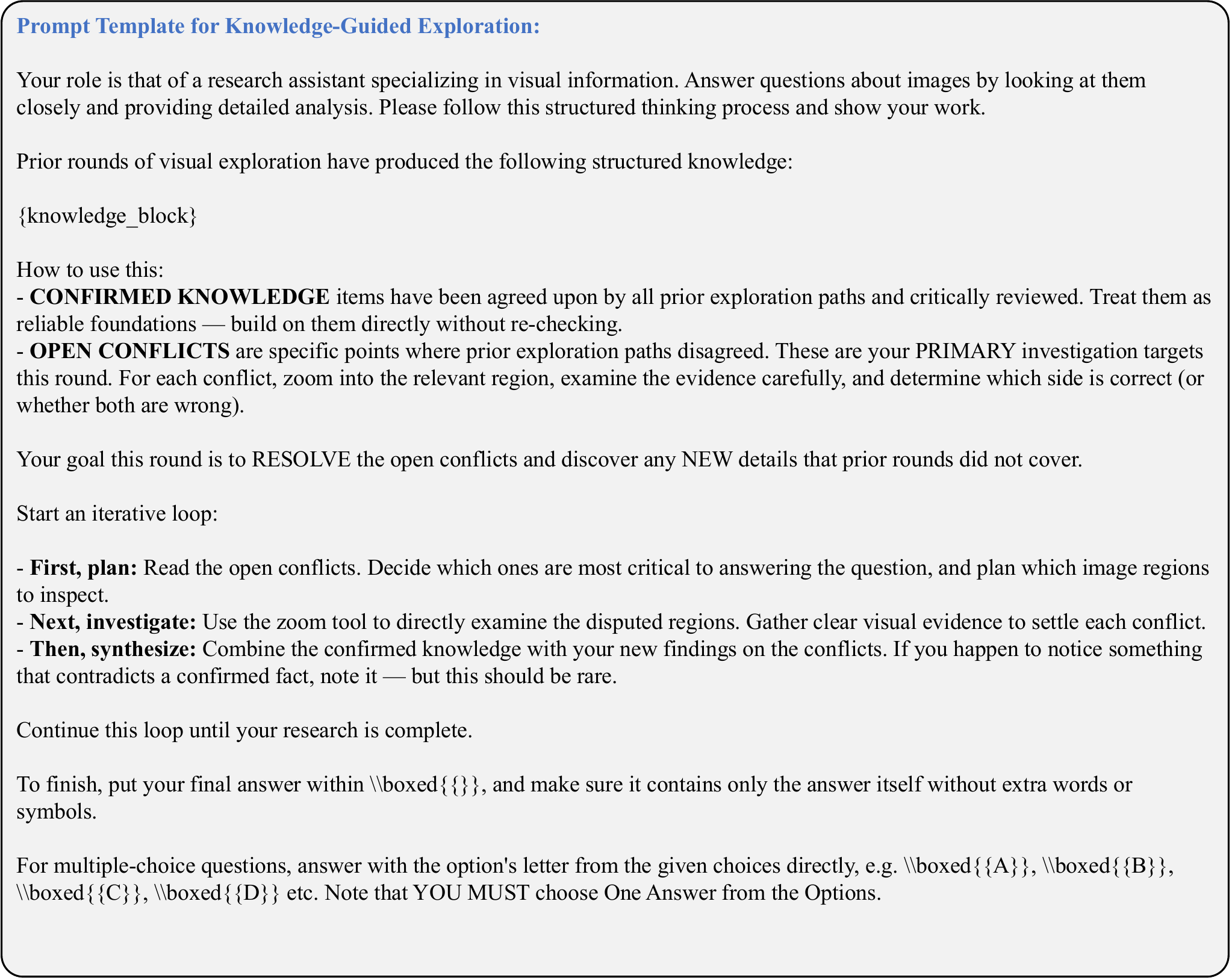}
  \caption{Prompt template for ledger-guided exploration.}
  \label{fig:knowledge_guided_exploration}
\end{figure*}

\begin{figure*}[t]
  \centering
  \includegraphics[width=0.9\linewidth]{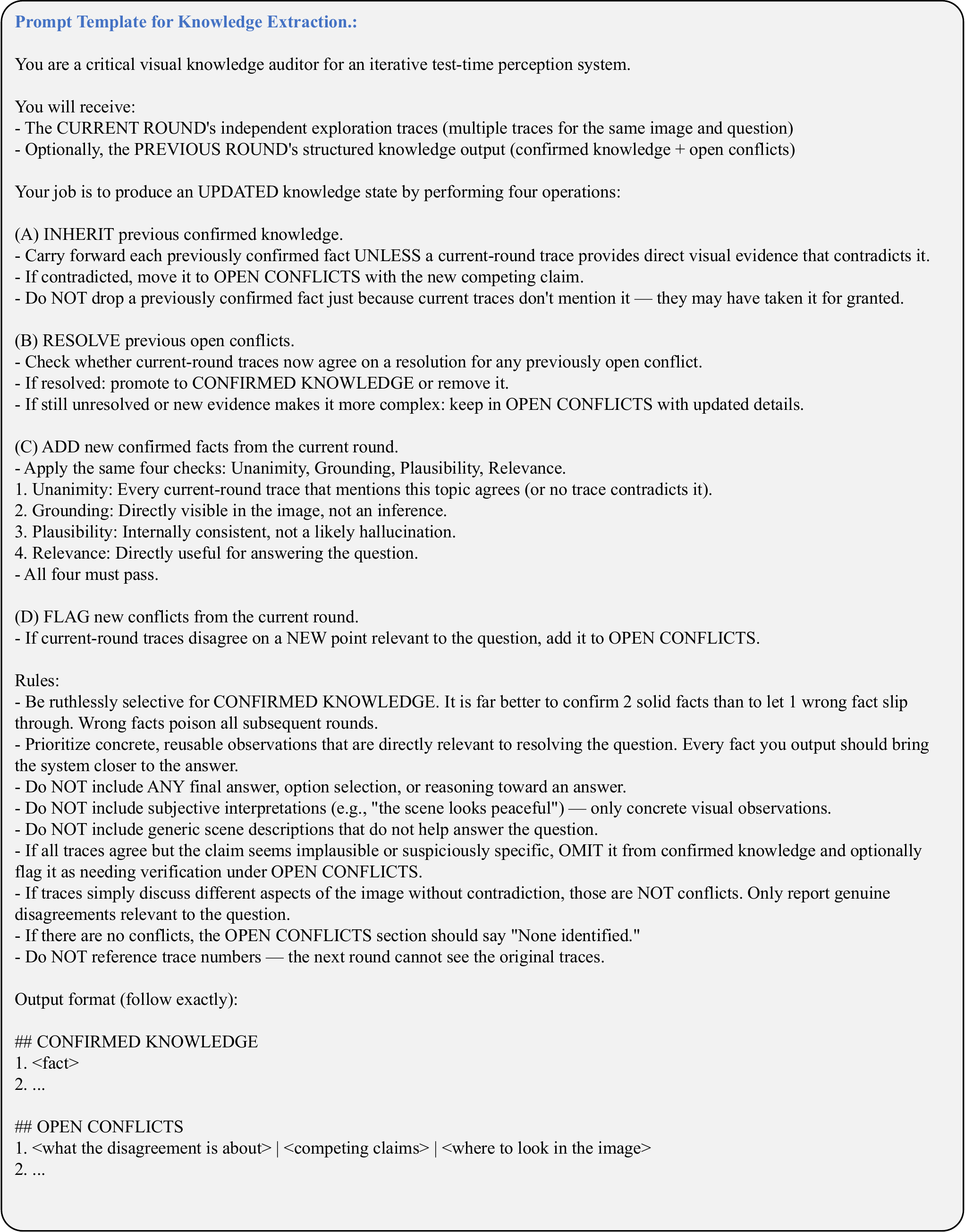}
  \caption{Prompt template for the Evidence Ledger update.}
  \label{fig:knowledge_extraction}
\end{figure*}

\begin{figure*}[t]
  \centering
  \includegraphics[width=0.9\linewidth]{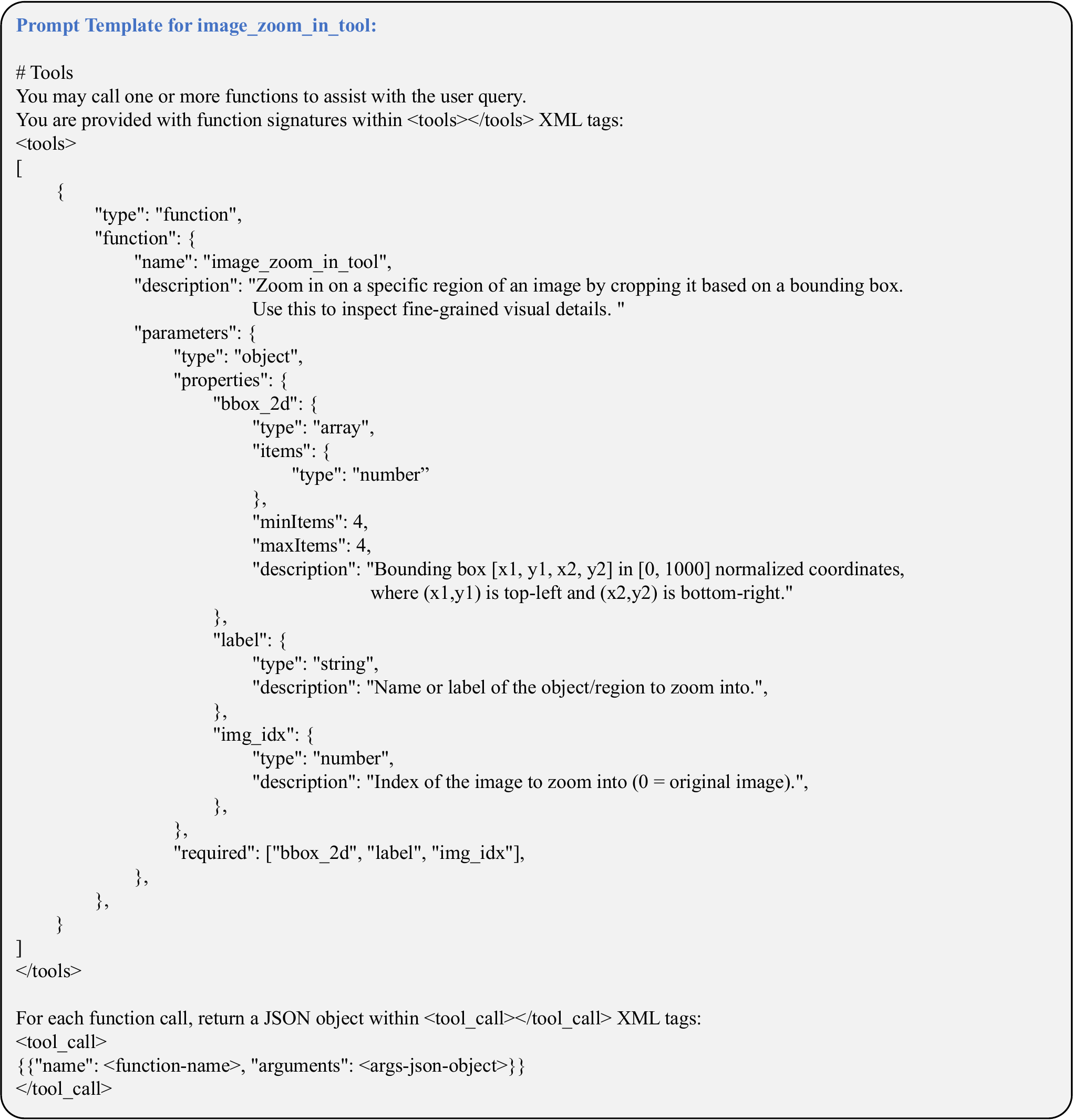}
  \caption{Specification of the crop tool exposed to the model.}
  \label{fig:tool}
\end{figure*}

The prompts implement the three operations of
Section~\ref{sec:methods}: trace sampling, the ledger update, and ledger-guided
sampling in later rounds. Exact templates appear in
Figures~\ref{fig:perceptual_exploration} to~\ref{fig:tool}.

\paragraph{Perceptual exploration.}
The sampling prompt casts the model as a visual assistant that answers an
image-grounded question by alternating between observing and inspecting. It asks
the model to state what is already legible in the current view, to invoke the
crop tool on the region it believes carries the missing detail, and to review the
returned view before deciding the next action. For standardized extraction, the
final response must be wrapped in \texttt{\textbackslash boxed\{\}}, and
multiple-choice tasks require only the option letter. The crop tool is exposed as
a function call taking a normalized bounding box, a textual region label, and an
image index, which keeps the calling format identical across benchmarks and makes
the inspected regions directly comparable with the annotations used in
Appendix~\ref{sec:grounding_rounds}.

\paragraph{Ledger update.}
After gating, the model acts as an auditor of visual evidence. It receives the
retained traces of the round and the previous ledger when one exists, and returns
the two updated tiers. The prompt states the four rules of
Section~\ref{sec:refinement} explicitly, namely retaining uncontradicted
entries, demoting contradicted ones, promoting conflicts that the round has
settled, and confirming a new claim only when it is consistent across independent
traces, visually supported, compatible with the existing confirmed tier, and
relevant to the question. Admission is deliberately conservative, since an
omission costs one further round of inspection whereas a false confirmation
propagates to every remaining round. Each open conflict must record the competing
claims together with a spatial directive naming the region to inspect next, which
is what makes the tier actionable rather than merely descriptive.

\paragraph{Ledger-guided sampling.}
From the second round onward the ledger is prepended to the system prompt of the
guided pool only. Confirmed Knowledge is presented as verified and not to be
re-checked, which frees crops for regions that remain unexplored, and Open
Conflicts are presented as prioritized targets together with their directives.
Fresh traces receive no ledger and therefore preserve an independent exploration
channel. This prompt-level separation is what realizes the exploration and
exploitation balance controlled by $\alpha$, and it is also why the ablation in
which the ledger is removed can be implemented simply by setting $\alpha=1$.

\section{Limitations and Future Work}
\label{app:limitations}

We group the limitations by the component they concern, so that each one points
at a specific part of the procedure rather than at the approach as a whole.

\paragraph{Selection.}
The gate reads a signal internal to a single trajectory and therefore cannot
detect an error that the model states confidently, a property visible in
Section~\ref{sec:analysis} as a correlation that is informative but far from
deterministic. Our design accommodates this by pairing the gate with
cross-trajectory consensus inside the ledger rather than by treating entropy as a
verifier, and Appendix~\ref{sec:reliability_measures} indicates that no
alternative intrinsic signal removes the issue. A learned verifier that reads the
inspected crops rather than only the token distribution would be a natural
replacement, and would additionally allow the gate to reject crops that are
visually irrelevant even when the accompanying text is stated with low entropy.

\paragraph{Utilization.}
The ledger is expressed in natural language and maintained by prompting a
deterministic summarization step. This keeps it readable, auditable, and easy to
correct, but it also makes its precision dependent on that step, and it leaves
each entry anchored to a region only through a textual reference. Representing
entries with explicit region identifiers, or learning the update rules instead of
prompting them, would tighten this link and would also allow the update to be
verified independently of the model that produced it.
A further caveat is that the ledger update is itself an image-conditioned model
call; although it is included in our token accounting, fully separating the
benefit of the ledger's dual-tier structure from that of the additional call
would require controls such as an image-free update or an unstructured summary
in its place, which we leave to future work.

\paragraph{Coverage and budget.}
\method\ remains a test-time scaling method and therefore costs more than
single-trajectory decoding, even though Section~\ref{sec:efficiency} shows that
it uses its budget more efficiently than competing multi-step baselines. It also
inherits the ability of the backbone to produce partially correct hypotheses,
since Eq.~\eqref{eq:coverage} improves with width only when the per-trace success
probability is non-zero. Both points suggest allocating the budget adaptively,
for instance by setting width, depth, and gate ratio per example according to an
estimate of how contested the evidence currently is, rather than fixing them in
advance.

\paragraph{Scope.}
Our study considers single images and a single visual operation. Extending the
ledger to multiple images, to video, and to embodied settings would test whether
a shared record of evidence remains tractable when observations must also be
indexed by time or by viewpoint, and enriching the operator set beyond cropping
would let the same machinery arbitrate among qualitatively different ways of
acquiring evidence.


\end{document}